\documentclass[10pt,twocolumn,letterpaper]{article}

\usepackage{cvpr}              

\newcommand{\cellc}{\cellcolor{lightgray!20}}

\newcommand{\softvq}{SoftVQ-VAE\xspace}
\newcommand{\vq}{VQ-VAE\xspace}
\newcommand{\kl}{KL-VAE\xspace}

\DeclareMathOperator*{\argmin}{arg\,min}
\DeclareMathOperator*{\softmax}{Softmax}

\definecolor{cvprblue}{rgb}{0.21,0.49,0.74}
\usepackage[pagebackref,breaklinks,colorlinks,allcolors=cvprblue]{hyperref}

\usepackage[capitalize]{cleveref}
\usepackage{nicefrac}       
\usepackage{microtype}      
\usepackage{xcolor}         
\usepackage{multirow}
\usepackage{colortbl}
\usepackage{wrapfig}
\usepackage{bbm}
\usepackage{enumitem}
\usepackage{booktabs}
\usepackage{graphicx}

\def\paperID{14217} 
\def\confName{CVPR}
\def\confYear{2025}

\title{SoftVQ-VAE: Efficient 1-Dimensional Continuous Tokenizer}

\author{
Hao Chen$^{1,2}$\thanks{Work done during internship at AMD. Email: haoc3@andrew.cmu.edu.},
Ze Wang$^{2}$,
Xiang Li$^{1}$,
Ximeng Sun$^{2}$,
Fangyi Chen$^{1}$,
Jiang Liu$^{2}$, \\
Jindong Wang$^{3}$,
Bhiksha Raj$^{1,4}$,
Zicheng Liu$^{2}$,
Emad Barsoum$^{2}$ \\
\small{$^{1}$Carnegie Mellon University, $^{2}$AMD,$^{3}$William \& Mary, $^{4}$MBZUAI}
}

\begin{document}
\maketitle
\begin{abstract}
Efficient image tokenization with high compression ratios remains a critical challenge for training generative models.
We present SoftVQ-VAE, a continuous image tokenizer that leverages soft categorical posteriors to aggregate multiple codewords into each latent token, substantially increasing the representation capacity of the latent space. 
When applied to Transformer-based architectures, our approach compresses 256$\times$256 and 512$\times$512 images using as few as 32 or 64 1-dimensional tokens.
Not only does SoftVQ-VAE show consistent and high-quality reconstruction, more importantly, it also achieves state-of-the-art and significantly faster image generation results across different denoising-based generative models. 
Remarkably, SoftVQ-VAE improves inference throughput by up to 18x for generating 256$\times$256 images and 55x for 512$\times$512 images while achieving competitive FID scores of 1.78 and 2.21 for SiT-XL.
It also improves the training efficiency of the generative models by reducing the number of training iterations by 2.3x while maintaining comparable performance. 
With its fully-differentiable design and semantic-rich latent space, our experiment demonstrates that SoftVQ-VAE achieves efficient tokenization without compromising generation quality, paving the way for more efficient generative models.
Code and model are released \footnote{\scriptsize{\url{https://github.com/Hhhhhhao/continuous_tokenizer}}.}.

\end{abstract}

\begin{figure*}[th!]
    \centering
    \includegraphics[width=0.95\linewidth]{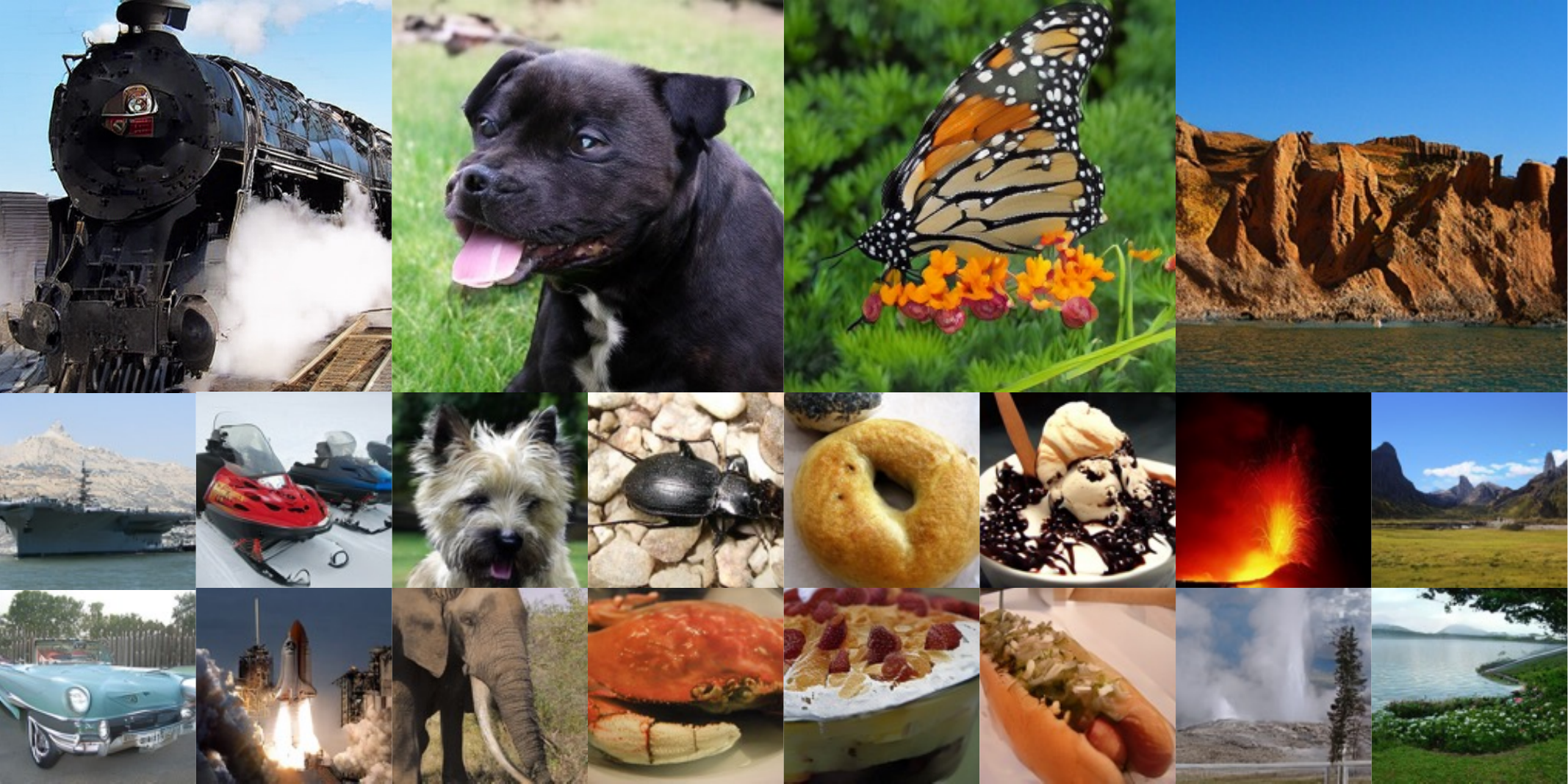}
    \vspace{-0.1in}
    \caption{ImageNet-1K 256$\times$256 and 512$\times$512 generation results of generative models trained on \softvq with 32 and 64 tokens. 
    }
    \label{fig:teaser}
    \vspace{-0.2in}
\end{figure*}

\section{Introduction}
\label{sec:intro}

\textit{Denoising}-based generative modeling has witnessed remarkable progress with recent advances, such as Diffusion Transformers (DiT) \cite{peebles2023scalablediffusionmodelstransformers}, Scalable Interpolant Transformers (SiT) \cite{ma2024sit}, and Masked Auto-Regressive models with diffusion loss (MAR) \cite{li2024autoregressiveimagegenerationvector}, to name a few. Denosing-based generative modeling not only presents impressive generation results on a wide range of modality, including natural language \cite{gong2022diffuseq,zheng2023reparameterized}, images \cite{ramesh2021zero,dhariwal2021diffusionmodelsbeatgans,rombach2022highresolutionimagesynthesislatent,lu2024simplifying,fan2024fluid}, videos \cite{ho2022video,blattmann2023stable,bar2024lumiere}, and audios \cite{popov2021grad,evans2024fast}, but also shows potential to unify the understanding and generation capabilities across modalities in multi-modal language models \cite{zhou2024transfusion,xie2024show,wang2024emu3,wu2024vila}.

A core component of denoising-based generative models is the tokenizer \cite{kingma2013auto,higgins2017beta,van2017neural,razavi2019generating,esser2021taming,zeghidour2021soundstream,lee2022autoregressiveimagegenerationusing,yu2023language}, which compresses the raw data of each modality into a set of 
latent tokens in either a \textit{discrete} or \textit{continuous} latent space.
The compact latent space therefore allows for more efficient and better generative modeling \cite{rombach2022highresolutionimagesynthesislatent}. 
Among previous efforts, Kullback–Leibler Variational Auto-Encoders (\kl) \cite{kingma2013auto} and Vector Quantized Variational Auto-Encoders (\vq) \cite{van2017neural,razavi2019generating,esser2021taming} stand out as representatives of tokenizers which introduce continuous and discrete latent spaces, respectively.
The former constrains the latent space with a Gaussian distribution using \textit{re-parametrization} \cite{kingma2013auto} trick, and the latter makes its latent space a categorical discrete distribution with a codebook of finite vocabulary which requires \textit{straight-through} estimation \cite{bengio2013estimating}.



Although both \kl and \vq (and their variants) have been predominantly adopted in many generative models \cite{rombach2022highresolutionimagesynthesislatent,peebles2023scalablediffusionmodelstransformers,chang2022maskgitmaskedgenerativeimage,sun2024autoregressive,zhou2024transfusion}, they still present two major challenges restricting the efficiency and effectiveness of generative modeling: (1) \textbf{the difficulty of achieving a higher compression ratio} \cite{chen2016variational} and 
(2) \textbf{the worse discriminative representations in their latent space than other self-supervised methods} \cite{he2022masked,chen2024deconstructing,yu2024representation,wu2024vila}.
The efficiency of downstream generative models, particularly Transformer-based architectures \cite{vaswani2023attentionneed}, is fundamentally constrained by their quadratic complexity to the number of latent tokens. Current image tokenizers \cite{esser2021taming,rombach2022highresolutionimagesynthesislatent,stabilityai2023} typically compress 256$\times$256 images to at least 256 tokens and 512$\times$512 images to at least 1024 tokens, creating a significant computational bottleneck for both training and inference of generative models \cite{peebles2023scalablediffusionmodelstransformers,ma2024sit,li2024autoregressiveimagegenerationvector}.
Many efforts have been made to reduce the number of latent tokens on the generative model side, such as merging \cite{peebles2023scalablediffusionmodelstransformers,ma2024sit,li2024imagefolder}, pooling \cite{li2023blip,song2024less,li2024tokenpacker}, and others \cite{luo2024feast,li2024mini,chen2024image}.
More studies have recently emerged to fundamentally reduce the token number of the tokenizer \cite{ge2023plantingseedvisionlarge,yu2024an,li2024imagefolder,chen2024deep}.
For example, TiTok \cite{yu2024an} uses 128 tokens, achieving generation results comparable to 256 tokens by adding one extra decoder.
DC-AE \cite{chen2024deep} compresses the initial 1024 tokens of 512$\times$512 with 256 tokens.
However, further increasing the compression ratio results in a significant degradation in the quality of reconstruction and therefore the generation \cite{chen2024deep}. 




Beyond the compression challenge, the quality of the learned representations of the tokenizer poses another critical limitation.
While latent representations are crucial for both understanding \cite{ge2023making,wu2024vila} and generation tasks \cite{ge2023plantingseedvisionlarge,lin2023sphinx,yu2024representation,li2024imagefolder}, current tokenizers usually struggle to capture more discriminative features.
Earlier methods attempted to advance the training objectives and recipe to improve the latent space of the tokenizer \cite{esser2021taming,dong2023peco,gu2023rethinkingobjectivesvectorquantizedtokenizers,zhu2024scaling,chen2024deconstructing}.
Recently, \cite{ge2023plantingseedvisionlarge,zhang2024codebook,yu2024spae,zhu2024scaling,wu2024vila,li2024imagefolder,yu2024representation} has also explored the representation alignment of latent tokens.
However, due to the lossy nature \cite{shannon1949communication,nyquist1928certain} of the smoothness constraints in \kl and the discrete quantization in \vq \cite{higgins2017beta,chen2016variational,zhu2024addressing,fifty2024restructuring}, current tokenizers usually fall short in learning the semantics of the latent space \cite{wu2024vila}.


To address the aforementioned challenges, we present \textbf{\softvq}, a simple modification to \vq that converts it from a discrete tokenizer into a continuous one with a high compression ratio and a semantic-rich latent space. 
Specifically, we propose using a \textbf{soft} categorical posterior in VAE with a learnable codebook. 
This straightforward modification introduces a crucial capability: instead of the conventional one-to-one mapping between tokens and codewords in \vq, our approach enables the adaptive aggregation of multiple codewords into each latent token, substantially increasing the representation capacity of the latent space.
We apply \softvq to the Transformer auto-encoder architecture \cite{yu2021vector,ge2023plantingseedvisionlarge,li2023blip,yu2024an,li2024imagefolder}, and we succeed in using much fewer 1-dimensional latent tokens (\ie, \textbf{32} and \textbf{64}) for both the reconstruction and subsequent generation tasks.
Since \softvq is now fully differentiable, it not only simplifies the learning of the codebook, not requiring the codebook loss or commit loss as in \vq \cite{van2017neural,esser2021taming}, but also enables better representation learning through direct alignment with pre-trained semantic features using a simple cosine similarity objective. 
Our approach generalizes the K-Means \cite{lloyd1982least} latent space of \vq to Soft K-Means~\cite{dunn1973fuzzy} and further to Gaussian Mixture Models (GMM) \cite{reynolds2009gaussian}, while maintaining compatibility with existing \vq techniques \cite{jegou2010product,yu2023language,lee2022autoregressive,li2024imagefolder}.
We comprehensively demonstrate the efficiency and effectiveness of the proposed \softvq in image reconstruction and generation tasks, and show that:
\begin{itemize}[leftmargin=1em,nosep]
\setlength\itemsep{0em}
    \item \softvq presents more consistent, robust, and high-quality reconstruction and generation results with various latent token lengths ranging from 256 to 32. 
    \item \softvq allows \textit{diffusion-based} DiT \cite{mentzer2023finite}, \textit{flow-based} SiT \cite{ma2024sit}, and \textit{autoregressive-based} MAR \cite{li2024autoregressiveimagegenerationvector} to achieve state-of-the-art generation results with only 64 and 32 tokens on both the 256$\times$256, with an FID of \textbf{1.78}, and 512$\times$512 on ImageNet benchmark, with an FID of \textbf{2.21}.
   \item  \softvq improves the inference throughput by \textbf{18}x and \textbf{10}x with 32 and 64 tokens to generate 256$\times$256 images, and \textbf{55}x with 64 tokens for generating 512$\times$512 images, with significantly lower GFLOPs for all models. 
   \item \softvq presents a more discriminative latent space through representation alignment. It also facilitates training efficiency by significantly improving training throughput and reduces training iterations by \textbf{2.3}x. 
\end{itemize}

\section{Preliminary}
\label{sec:prelimiary}

We present an overview of the tokenizers, \ie, \kl \cite{kingma2013auto} and \vq \cite{van2017neural}, and denoising-based generative models, \ie, DiT \cite{peebles2023scalablediffusionmodelstransformers}, SiT \cite{ma2024sit}, and MAR \cite{li2024autoregressiveimagegenerationvector}, in this section.

\subsection{Image Tokenizer}

The architecture of image tokenizer generally resembles auto-encoders, consisting of two main components: an encoder $\mathcal{E}$ and a decoder $\mathcal{D}$, parameterized by $\phi$ and $\theta$, respectively.
Given an input image $\mathbf{x} \in \mathbb{R}^{H \times W \times 3}$, the encoder $\mathcal{E}$ maps the high-dimensional $\mathbf{x}$ to a lower-dimensional latent representation $\mathbf{z} = \mathcal{E}(\mathbf{x};\phi)$.
The latent representations can have varying shapes depending on the encoder architecture. 
We generally view the latent representation as a set of latent tokens $\mathbf{z} = [\mathbf{z}^{[0]}, \mathbf{z}^{[1]}, \dots, \mathbf{z}^{[L]}] \in \mathbb{R}^{L \times D}$, where $L$ is the number of tokens and $D$ is the dimension of the latent representation.
For basic auto-encoders (AE), the decoder takes latent tokens as input and reconstructs the original signal $\hat{\mathbf{x}} = \mathcal{D}(\mathbf{z};\theta)$ by minimizing the reconstruction loss with $\mathbf{x}$.

Variational auto-encoders (VAE) \cite{kingma2013auto} extend the basic AE by introducing a probabilistic perspective on the latent space. 
Specifically, VAE approximates a posterior distribution $p(\mathbf{z} | \mathbf{x})$ with a learned distribution $q_{\phi}(\mathbf{z}|\mathbf{x})$ from the encoder, assuming that $\mathbf{x}$ is generated by the unobserved latent $\mathbf{z}$ with a prior distribution $p(\mathbf{z})$. 
The decoder instead takes a set of sampled latent tokens $\mathbf{z} \sim q_{\phi}(\mathbf{z}|\mathbf{x})$ from the posterior and aims to learn the marginal likelihood of data in a generative process by optimizing the evidence lower bound (ELBO):
$\max_{\phi, \theta} \mathbb{E}_{q_{\phi}(\mathbf{z} | \mathbf{x})} \log p_{\theta}(\mathbf{x} | \mathbf{z}) - \mathcal{L}_{\mathrm{kl}}(  q_{\phi} (\mathbf{z} | \mathbf{x}) | p(\mathbf{z}) ) $. 
To make the optimization tractable, assumptions have been made on the prior, leading to different variants of VAE. 

\noindent \textbf{KL-VAE} \cite{kingma2013auto,higgins2017beta} parametrizes both the prior and posterior as Gaussians.
The prior $p(\mathbf{z})$ is assumed to be the isotropic unit Gaussian $\mathcal{N}(0, 1)$. 
The \textit{continuous} latent code is parameterized with posterior mean $\boldsymbol{\mu}_{\phi}$ and variance $\boldsymbol{\sigma}^2_{\phi}$ predicted by the encoder using the ``re-parametrization" trick with a noise variable $\boldsymbol{\varepsilon}$ from standard Gaussian:
\begin{equation}
\resizebox{.8\linewidth}{!}{$
\begin{split}
    & \textit{posterior:  } q_{\phi}(\mathbf{z} | \mathbf{x}) = \mathcal{N}(\mathbf{z}; \boldsymbol{\mu}_{\phi}(\mathbf{x}), \boldsymbol{\sigma}^2_{\phi}(\mathbf{x})) \\
    & \textit{latent:  } \mathbf{z} = \boldsymbol{\mu}_{\phi}(\mathbf{x}) + \boldsymbol{\sigma}_{\phi}(\mathbf{x}) \odot \mathbf{\varepsilon}, \quad \epsilon \sim \mathcal{N}(0, \mathbf{I}) \\
    & \textit{kl: } \mathcal{L}_{\mathrm{kl}} = -\frac{1}{2}\left(1+\log \left(\boldsymbol{\sigma}_{\phi}^2(\mathbf{x})\right)-\boldsymbol{\mu}_{\phi}(\mathbf{x})^2-\boldsymbol{\sigma}_{\phi}^2(\mathbf{x})\right)
\end{split}
$}
\label{eq:kl}
\end{equation}
While KL-VAE is widely used in diffusion models, it is known to have the ``posterior-collapse" issue \cite{chen2016variational}. 
In addition, the KL loss weight usually imposes a trade-off between the quality of reconstruction and the smoothness of its latent space \cite{tschannen2025givt}, preventing high compression ratio and the learning of semantics in the latent space of \kl.

\noindent \textbf{VQ-VAE} \cite{van2017neural,esser2021taming}, in contrast, generates latent code in a \textit{discrete} space by taking the posterior as a $K$-way deterministic categorical distribution against a learnable codebook $\mathcal{C} = [\mathbf{c}^{[0]}, \dots, \mathbf{c}^{[K]}] \in \mathbb{R}^{K \times D}$, with $\mathbf{c}$ as the codeword:
\begin{equation}
\resizebox{.9\linewidth}{!}{$
\begin{split}
  & \textit{posterior: } q_{\phi}(\mathbf{z} = k | \mathbf{x}) = 
   \begin{cases} 
      1 & \text{if } k = \argmin_{j} \| \hat{\mathbf{z}} - \mathbf{c}^{[j]} \|_2 \\
      0 & \text{otherwise} 
   \end{cases} \\
  & \textit{latent: } \mathbf{z} = \mathbf{c}^{[k]}, \quad \text{where } k = \argmin_{j} \| \hat{\mathbf{z}} - \mathbf{c}^{[j]} \|_2  \\ 
  & \textit{kl: } \mathcal{L}_{\mathrm{kl}} = \log K  \\
\end{split}
$}
\label{eq:vq}
\end{equation}
Due to the non-differentiable $\argmin$, a ``straight-through'' trick \cite{bengio2013estimating} is adopted to approximate the gradient of the encoder output $\hat{\mathbf{z}}$ by directly copying gradient from the decoder input $\mathbf{z}$. 
Since the reconstruction objective does not impose direct gradients on the codebook $\mathcal{C}$, VQ-VAE additionally uses a codebook loss as $\|\text{sg}[\hat{\mathbf{z}}] - \mathbf{c}\|_2$ to move codewords toward the encoder output, along with a commit loss $\|\hat{\mathbf{z}} - \text{sg}[\mathbf{c}]\|_2$ to prevent arbitrary growth of codewords, where $\text{sg}[\cdot]$ denotes stop-gradient.
The broken gradient hinders high compression ratio and latent space learning \cite{huh2023straightening,fifty2024restructuring}.



\subsection{Denoising-based Generative Models}

Denoising-based generative models synthesize images by progressively transforming Gaussian noise $\boldsymbol{\varepsilon} \sim \mathcal{N}(0, \mathbf{I})$ into tokenizer latent codes $\mathbf{z}$ through a forward process \cite{sohl2015deep,ho2020denoising}:
\begin{equation}
\mathbf{z}_t = \alpha_t \mathbf{z} + \sigma_t \varepsilon,
\label{eq:forward}
\end{equation}
where $\alpha_t$ and $\sigma_t$ are a decreasing and increasing function of $t$, and $t \in [0, T]$ \cite{kingma2024understanding}. In this paper, we adopt three approaches to implementing this denoising process as follows.

\noindent \textbf{DiT} \cite{peebles2023scalablediffusionmodelstransformers} is a diffusion-based model with a transformer architecture. 
It formulates the process through a forward-time stochastic differential equation (SDE), where $\mathbf{z}_t$ converges to $\mathcal{N}(0, \mathbf{I})$ as $t \rightarrow T$. 
Generation occurs via a reverse-time SDE \cite{anderson1982reverse}, with the model trained to predict noise $\boldsymbol{\varepsilon}$ at randomly sampled timesteps. 
Due to the long token length of the tokenizer \cite{stabilityai2023} used in DiT, it has several variants to merge tokens at the input of the transformer backbone.



\noindent \textbf{SiT} \cite{ma2024sit} instead uses stochastic interpolants \cite{albergo2023stochastic} and performs denoising via probability flow ordinary differential equation (PF ODE). 
The model learns the ODE's velocity field \cite{lipman2022flow} by minimizing mean-squared-error (MSE) loss.
This velocity field defines a score function for generation, analogous to DiT's reverse-time SDE.


\noindent \textbf{MAR} \cite{li2024autoregressiveimagegenerationvector} 
combines diffusion with autoregressive generation. 
Unlike DiT and SiT's parallel denoising of all latent tokens, MAR adopts an encoder-decoder transformer architecture \cite{he2022masked} that progressively denoises tokens in a ``next-set'' autoregressive manner \cite{tian2024visualautoregressivemodelingscalable}, starting from masked tokens, similar to MaskGIT \cite{chang2022maskgitmaskedgenerativeimage} and MAGE \cite{li2023magemaskedgenerativeencoder}.


\section{Method}
\label{sec:method}

In this section, we present \softvq, a novel 1D continuous tokenizer with a high compression ratio and a semantic-rich latent space. 
We first introduce the architecture and the formulation of \softvq. 
Then, we show that \softvq can learn semantics easily by aligning its latent tokens with pre-trained features via its fully-differentiable property.



\subsection{Architecture}

\begin{figure}[t!]
    \centering
    \includegraphics[width=0.98\linewidth]{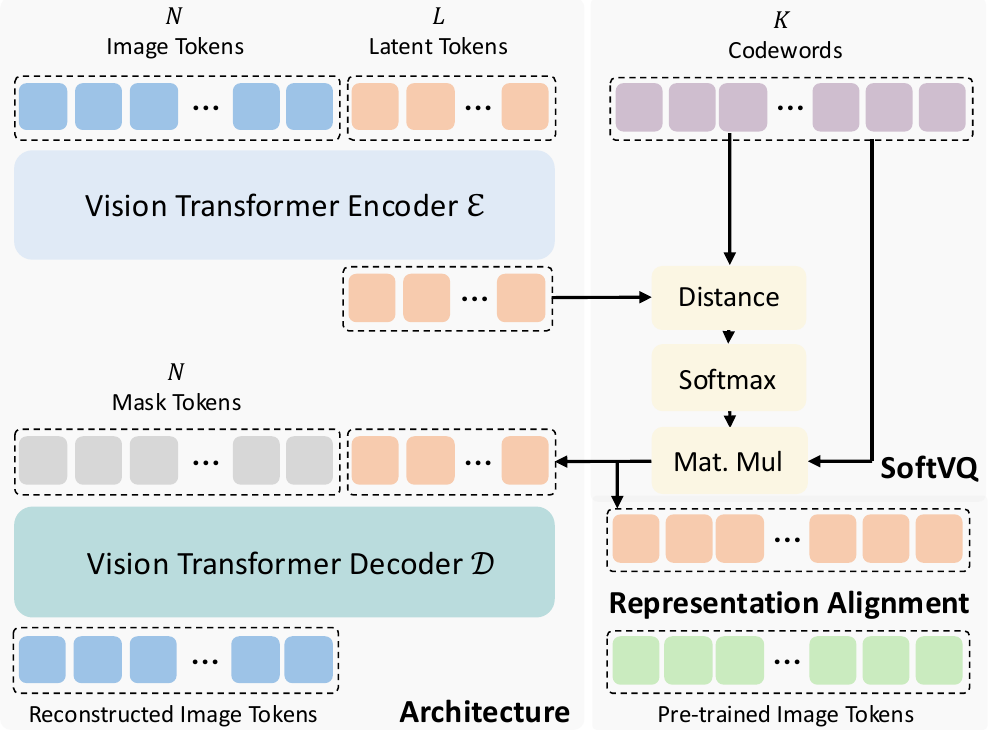}
    \vspace{-0.1in}
    \caption{Illustration of \softvq. Left: Transformer encoder-decoder architecture with image tokens, arbitrary length of latent tokens, and mask tokens. Right top: fully-differentiable SoftVQ illustration. Right bottom: latent space representation alignment.}
    \label{fig:architecture}
\vspace{-0.2in}
\end{figure}

We leverage Vision Transformer (ViT) \cite{dosovitskiy2021imageworth16x16words,yu2021vector} as the architecture for the encoder $\mathcal{E}$ and decoder $\mathcal{D}$ of \softvq.
Similarly to \citet{yu2024an}, instead of using the image features as latent, we initialize a set of extra 1D learnable tokens and use these tokens for reconstruction and subsequent generation.
With the self-attention mechanism, it allows the learnable tokens to adaptively aggregate different image tokens to obtain the latent tokens of SoftVQ, detailed in \cref{sec:method-softvq}. 
Specifically, for the encoder transformer with a patch size of $P$, an image $\mathbf{x} \in \mathbb{R}^{H \times W \times 3}$ is patchified into a sequence of image tokens $\mathbf{x}_p \in \mathbb{R}^{N \times D}$, where $N = \ HW / P^2$ and $D$ represent the token dimension size.
We then concatenate a set of learnable tokens $\mathbf{z}_l \in \mathbb{R}^{L \times D}$ of length $L$ along with the image tokens as the input to the encoder, and retain only the output corresponding to the learnable latent tokens as the output of our encoder: $\hat{\mathbf{z}} = \mathcal{E}([\mathbf{x}_p; \mathbf{z}_l];\phi)$. 

For the decoder, we reconstruct images from a sequence of learnable mask tokens $\mathbf{m} \in \mathbb{R}^{N \times D}$ \cite{bao2021beit,he2022masked}, concatenated with latent tokens: $\hat{\mathbf{x}} = \mathcal{D}([\mathbf{m};\mathbf{z}];\theta)$. 
We use a linear layer at the end of the decoder to regress the pixel values from the mask tokens.
For the image tokens at the encoder input and mask tokens at the decoder input, we apply 2D absolute position embeddings and 1D absolute position embedding for the latent tokens. 
Adopting more advanced position embedding techniques, such as RoPE \cite{su2024roformer,heo2025rotary}, may lead to better results and is left for future work. 
This design allows \softvq to use latent codes of arbitrary length, each adaptively associated with image tokens, for reconstruction and subsequent generative modeling, and model images of various resolutions using the same length of latent codes.
An overview of the model architecture is shown in \cref{fig:architecture}.


\subsection{\softvq}
\label{sec:method-softvq}

As discussed above, the high compression ratio of both KL-VAE and VQ-VAE usually results in a significant degradation of quality in reconstruction and latent space.
To overcome these limitations, we propose \softvq, a simple modification to VQ-VAE, which maintains the advantages of learning codewords to capture the data distribution while bridging it with more representation capacity as a continuous tokenizer. 
The core idea of \softvq is to allow each latent code to adaptively aggregate multiple codewords from the learnable codebook. 
Consequently, it imposes a \textit{soft} categorical distribution on the posterior from the encoder:
\begin{equation}
\resizebox{.75\linewidth}{!}{$
\begin{split}
    & \textit{posterior: }  q_{\phi}(\mathbf{z} | \mathbf{x}) = \softmax \left(-\frac{\| \hat{\mathbf{z}} - \mathcal{C} \|_2}{\tau}\right) \\
    & \textit{latent: } \mathbf{z} = q_{\phi}(\mathbf{z} | \mathbf{x}) \mathcal{C} \\ 
    & \textit{kl: } \mathcal{L}_{\mathrm{kl}} = H(q_{\phi}(\mathbf{z} | \mathbf{x}))- H\left(\mathbb{E}_{\mathbf{x} \sim p(\mathbf{x})} q_{\phi}(\mathbf{z} | \mathbf{x})\right) \\ 
\end{split}
$}
\end{equation}
where $\tau$ is the temperature parameter controlling the sharpness of the softmax probability and is set to $0.07$ by ablation. 
The derivation of $\mathcal{L}_{kl}$ is shown in \cref{sec:appendix-softvq-kl}.

Despite this simple modification, \softvq can significantly reduce the latent token length while maintaining a high quality of the reconstruction and latent space. 
More importantly, with a soft categorical distribution, \softvq is fully-differentiable, and thus the encoder and codebook can be optimized directly from the reconstruction loss (and other losses). 
It also allows for easier regularization of various forms on the latent space that drastically improves its quality, and less hyper-parameter tuning without codebook and commit loss \cite{huh2023straightening} in the reduced training objectives.

\noindent \textbf{Interpretation}. 
The posterior and latent space of \vq can be interpreted with K-Means \cite{lloyd1982least}.
Hence, \softvq can be viewed as soft K-Means \cite{dunn1973fuzzy}, with codewords as learnable prototypes in the latent space.
This latent space can easily be extended to Gaussian Mixture Models (GMM) \cite{reynolds2009gaussian}, which we refer to as \textbf{GMMVQ-VAE}.
Instead of using fixed temperature, GMMVQ-VAE predicts data-dependent weights for the codeword prototypes, \ie, Gaussian means, to compute the soft categorical posterior: $q_{\phi}(\mathbf{z}|\mathbf{x}) = \softmax(- \omega(\hat{\mathbf{z}})  \| \hat{\mathbf{z}} - \mathcal{C} \|_2 )$.
Although prior work has explored GMM-based VAEs \cite{dilokthanakul2016deep,johnson2016composing}, our GMMVQ-VAE differs by learning discrete prototypes rather than continuous latent variables.
While providing the additional benefit of interpretable Gaussian components in the latent space, GMMVQ achieves reconstruction quality and downstream generation performance similar to SoftVQ in our experiments, and thus we mainly adopt the simpler SoftVQ throughout this paper.
More details are in \cref{sec:appendix-softvq-gmm}.

\subsection{Representation Alignment of Latent Space}

While the reconstruction quality of tokenizer is important, learning a high-quality latent space is more crucial for downstream denoising-based generative modeling \cite{yu2024representation}.
Several recent approaches have explored aligning the latent code with pre-trained features through a contrastive loss \cite{ge2023plantingseedvisionlarge,wu2024vila,li2024imagefolder} and initializing the codebook with pre-trained features \cite{zhu2024scaling}. 
However, imposing effective regularization on the latent remains a challenge due to the discrete quantization of \vq and the Gaussian constraints in \kl. 

Thanks to the fully-differentiable property of \softvq, we can now directly impose regularization on the latent space.
Inspired by the recent work of REPA \cite{yu2024representation}, we propose aligning the representations of latent codes with pre-trained vision encoders to facilitate learning the latent space.
Compared to REPA, which aligns the representation at intermediate layers of downstream generative models, our method of feature alignment in the tokenizer's latent space is equivalent to performing REPA at the input space of generative models. 
Specifically, to align the latent code with the image tokens from a pre-trained vision encoder, we replicate each latent token by $N / L$ times:
\begin{equation}
    \mathbf{z}_r = [\underbrace{\mathbf{z}^{[0]}, \ldots, \mathbf{z}^{[0]}}_{N / L \text{ times}}, \underbrace{\mathbf{z}^{[1]}, \ldots, \mathbf{z}^{[1]}}_{N / L \text{ times}}, \ldots, \underbrace{\mathbf{z}^{[L]}, \ldots, \mathbf{z}^{[L]}}_{N / L \text{ times}}],
\end{equation}
and encourage the token-wise similarity with image tokens.
We employ a projector multilayer perceptron (MLP) on top of the latent tokens to match the pre-trained feature $\mathbf{y}_*$:
\begin{equation}
\mathcal{L}_{\mathrm{align}} = \frac{1}{N} \sum_{n=1}^N \operatorname{sim}\left(\mathbf{y}_*^{[n]}, \text{MLP}\left(\mathbf{z}_r^{([n]}\right)\right).
\end{equation}

This alignment ensures the latent space of \softvq captures semantically discriminative features that are beneficial for training the downstream generative models, even if it does not directly translate into improved reconstruction performance from the tokenizer. 
However, as we will show in \cref{sec:exp-latent}, the generation quality of trained generative models often depends more on the semantic structure of the latent space than on the tokenizer's reconstruction capabilities.

\subsection{Final Training Objective}

The training objective of \softvq combines the reconstruction loss, perceptual loss \cite{larsen2016autoencoding,johnson2016perceptual,dosovitskiy2016generating,zhang2018unreasonableeffectivenessdeepfeatures} and adversarial loss \cite{goodfellow2020generative,isola2018imagetoimagetranslationconditionaladversarial} as in VQ \cite{esser2021taming}, and representation alignment:
\begin{equation}
\mathcal{L} = \mathcal{L}_{\textrm{recon}} + \lambda_1 \mathcal{L}_{\textrm{percep}} + \lambda_2 \mathcal{L}_{\textrm{adv}} + \lambda_3 \mathcal{L}_{\textrm{align}} + \lambda_4 \mathcal{L}_{\textrm{KL}},
\end{equation}
where $\lambda_1$, $\lambda_2$, $\lambda_3$, and $\lambda_4$ are hyper-parameters. 
Neither codebook nor commit loss as in VQ is needed.
The perceptual loss $\mathcal{L}_{\textrm{percep}}$ helps capture high-level perceptual features, $\mathcal{L}_{\textrm{adv}}$ encourages the decoder to generate realistic images by removing the artifacts from the reconstruction loss only, $\mathcal{L}_{\textrm{align}}$ guides the latent space to align with pre-trained features, and $\mathcal{L}_{\textrm{kl}}$ as the KL divergence term in the ELBO. 

\section{Experiments}
\label{sec:exp}

In this section, we validate the efficiency and effectiveness of \softvq with extensive experiments.

\begin{table*}[t!]
\centering
\caption{\textbf{System-level comparison} on ImageNet 256$\times$256 conditional generation. We compare with both diffusion-based models and auto-regressive models with different types of tokenizers. \colorbox{lightgray!20}{Grey} denotes SoftVQ. $^\dagger$ indicates results with DPM-Solver and 50 steps. }
\vspace{-0.1in}
\label{tab:main_256}
\resizebox{0.75\linewidth}{!}{%
\begin{tabular}{@{}lccc|cc|cc|cc@{}}
\toprule
\multirow{2}{*}{Gen. Model} &
  \multirow{2}{*}{Tok. Model} &
  \multirow{2}{*}{\# Tokens $\downarrow$} &
  \multirow{2}{*}{Tok. rFID  $\downarrow$} &
  \multirow{2}{*}{Gflops $\downarrow$} &
  \multirow{2}{*}{\begin{tabular}[c]{@{}c@{}}Throughput\\ (imgs/sec)\end{tabular}  $\uparrow$} &
  \multicolumn{2}{c|}{w/o CFG} &
  \multicolumn{2}{c}{w/ CFG} \\ 
 &
   &
   &
   &
   &
   &
  gFID  $\downarrow$ &
  IS  $\uparrow$ &
  gFID  $\downarrow$ &
  IS $\uparrow$ \\ \toprule

\multicolumn{10}{c}{\textbf{Auto-regressive Models}} \\ \bottomrule

Taming-Trans.  \cite{esser2021taming} &
  VQ &
  256 &
  7.94 &
  - &
  - &
  5.20 &
  290.3 &
  - &
  - \\
RQ-Trans.  \cite{lee2022autoregressive} &
  RQ &
  256 &
  3.20 &
  908.91 &
  7.85 &
  3.80 &
  323.7 &
  - &
  - \\
MaskGIT \cite{chang2022maskgitmaskedgenerativeimage} &
  VQ &
  256 &
  2.28 &
  - &
  - &
  6.18 &
  182.1 &
  - &
  - \\
MAGE \cite{li2023magemaskedgenerativeencoder} &
  VQ &
  256 &
  - &
  - &
  - &
  6.93 &
  195.8 &
  - &
  - \\
LlamaGen-3B \cite{sun2024autoregressive} &
  VQ &
  256 &
  2.19 &
  781.56 &
  2.90 &
  - &
  - &
  3.06 &
  279.7 \\
TiTok-S-128 \cite{yu2024an} &
  VQ &
  128 &
  1.61 &
  33.03 &
  6.50 &
  - &
  - &
  1.97 &
  281.8 \\
MAR-H \cite{li2024autoregressiveimagegenerationvector} &
  KL &
  256 &
  1.22 &
  145.08 &
  0.12 &
  2.35 &
  227.8 &
  1.55 &
  303.7 \\  \midrule
\cellc  &
  \cellc   SoftVQ-L &
\cellc   32 &
\cellc   0.61 &
\cellc   67.93 &
\cellc   2.08 &
\cellc   3.83 &
\cellc   211.2 &
\cellc   2.54 &
\cellc   273.6
\\
\multirow{-2}{*}{\cellc  MAR-H} &
  \cellc  SoftVQ-BL &
\cellc    64 &
\cellc   0.65 &
\cellc   86.55 &
\cellc   0.89 &
\cellc   2.81 &
\cellc   218.3 &
\cellc   1.93 &
\cellc   289.4 
\\
 \toprule
\multicolumn{10}{c}{\textbf{Diffusion-based Models}} \\ \bottomrule
LDM-4 \cite{rombach2022highresolutionimagesynthesislatent} &
  KL &
  4096 &
  0.27 &
  157.92 &
  0.37 &
  10.56 &
  103.5 &
  3.60 &
  247.7 \\
U-ViT-H/2$^\dagger$ \cite{bao2023all} &
  \multicolumn{1}{c}{\multirow{5}{*}{KL}} &
  \multirow{5}{*}{1024} &
  \multicolumn{1}{c|}{\multirow{5}{*}{0.62}} &
  128.89 &
  0.98 &
  - &
  - &
  2.29 &
  263.9 \\
MDTv2-XL/2 \cite{gao2023mdtv2} &
   &
   &
   &
  125.43 &
  0.59 &
  5.06 &
  155.6 &
  1.58 &
  314.7 \\
DiT-XL/2 \cite{peebles2023scalablediffusionmodelstransformers} &
   &
   &
   &
  80.73 &
  0.51 &
  9.62 &
  121.5 &
  2.27 &
  278.2 \\
SiT-XL/2 \cite{ma2024sit} &
   &
   &
   &
  \multirow{2}{*}{81.92} &
  \multirow{2}{*}{0.54} &
  8.30 &
  131.7 &
  2.06 &
  270.3 \\
+ REPA \cite{yu2024representation} &
   &
   &
   &
   &
   &
  5.90 &
  157.8 &
  1.42 &
  305.7 \\ \midrule
\cellc &
\cellc   SoftVQ-B &
\cellc &
\cellc   0.89 &
\cellc &
\cellc   8.94 &
\cellc   9.83 &
\cellc   113.8 &
\cellc   3.91  &
\cellc    264.2 \\
\cellc  &
\cellc   SoftVQ-BL &
\cellc    &
\cellc   0.68 &
\cellc    &
\cellc   8.81 &
\cellc   9.22 &
\cellc   115.8 &
\cellc   3.78  &
\cellc   266.7  \\
  \multirow{-3}{*}{\cellc DiT-XL} &
\cellc   SoftVQ-L &
   \multirow{-3}{*}{\cellc  32} &
\cellc    0.74 &
     \multirow{-3}{*}{\cellc 14.52} &
\cellc   8.74 &
\cellc   9.07 &
\cellc   117.2 &
\cellc   3.69 &
\cellc   270.4  \\   \midrule
\cellc   &
\cellc   SoftVQ-B &
\cellc   &
\cellc   0.88 &
\cellc   &
\cellc   4.70 &
\cellc   6.62  &
\cellc   129.2 &
\cellc   3.29  &
\cellc   262.5  \\
\cellc   &
\cellc   SoftVQ-BL &
 \cellc   &
\cellc   0.65 &
\cellc    &
\cellc   4.59 &
\cellc   6.53 &
\cellc   131.9 &
 \cellc  3.11 &
\cellc   268.3  \\
\multirow{-3}{*}{\cellc DiT-XL} &
\cellc   SoftVQ-L &
\multirow{-3}{*}{\cellc 64} &
\cellc   0.61 &
  \multirow{-3}{*}{\cellc  28.81} &
\cellc   4.51 &
\cellc   5.83 &
\cellc   141.3 &
\cellc   2.93  &
\cellc   268.5  \\ \midrule
\cellc  &
\cellc   SoftVQ-B &
\cellc    &
 \cellc  0.89 &
\cellc    &
\cellc   10.28 &
\cellc   7.99 &
\cellc   129.3 &
\cellc   2.51 &
\cellc   301.3 \\
\cellc  &
\cellc   SoftVQ-BL &
\cellc    &
\cellc   0.68 &
\cellc    &
\cellc   10.12 &
\cellc   7.67  &
\cellc   133.9 &
\cellc   2.44 &
\cellc   308.8 \\
 \multirow{-3}{*}{\cellc SiT-XL} &
\cellc   SoftVQ-L &
 \multirow{-3}{*}{\cellc   32} &
\cellc  0.74 &
\multirow{-3}{*}{\cellc 14.73}  &
\cellc   10.11 &
\cellc   7.59 &
\cellc   137.4 &
\cellc   2.44 &
\cellc   310.6 \\ \midrule
\cellc   &
\cellc    SoftVQ-B &
\cellc     &
\cellc    0.88 &
\cellc    &
\cellc    5.51 &
\cellc    5.98 &
\cellc    138.0 &
\cellc    1.78 &
 \cellc   279.0 \\
\cellc &
\cellc    SoftVQ-BL &
\cellc     &
\cellc    0.65 &
 \cellc    &
\cellc    5.42 &
\cellc    5.80 &
\cellc    143.5 &
\cellc    1.88 &
\cellc    287.9 \\
 \multirow{-3}{*}{\cellc SiT-XL} &
\cellc    SoftVQ-L &
\multirow{-3}{*}{\cellc 64}  &
\cellc    0.61 &
 \multirow{-3}{*}{\cellc 29.23}  &
\cellc    5.39 &
\cellc    5.35 &
\cellc    151.2 &
\cellc    1.86 &
\cellc    293.6 \\ 
 \bottomrule
\end{tabular}%
}
\vspace{-0.2in}
\end{table*}

\subsection{Experiments Setup}
\label{sec:exp-setup}

\noindent \textbf{Implementation Details of \softvq}. 
We use the LlamaGen codebase \cite{sun2024autoregressive} to build our \softvq. 
We instantiate 4 configurations: SoftVQ-S, SoftVQ-B, SoftVQ-BL, SoftVQ-L, with a total of 45M, 173M, 391M, and 608M parameters, respectively.
Each configuration has variants of latent codes $L=64$ and $L=32$.
For the representation alignment, we choose DINOv2 \cite{oquab2023dinov2} for the pre-trained features, similarly as \citet{yu2024representation} and \citet{li2024imagefolder}. 
To learn a better semantic latent space, we initialize our encoder from the pre-trained DINOv2 weights. 
Alignment with other pre-trained features is explored in \cref{sec:exp-latent}.
We train the tokenizers on ImageNet \cite{deng2009imagenet} of resolution 256$\times$256 and 512$\times$512 for 250K iterations. 
Similarly to \citet{tian2024visualautoregressivemodelingscalable} and \citet{li2024imagefolder}, we adopted the same frozen DINO-S  \cite{caron2021emerging,oquab2023dinov2} discriminator with a similar architecture to StyleGAN \cite{karras2019style,karras2020analyzing}.
We use DiffAug \cite{zhao2020differentiable}, consistency regularization \cite{zhang2019consistency}, and LeCAM regularization \cite{tseng2021regularizinggenerativeadversarialnetworks} for discriminator training as in \cite{tian2024visualautoregressivemodelingscalable}.
For the training objective, we set $\lambda_1=1.0$, $\lambda_1=0.2$, $\lambda_3=0.1$, and $\lambda_4=0.01$, following previous common practice. 
More training details are shown in \cref{sec:appendix-exp-softvq}.

\noindent \textbf{Implementation Details of Generative Modeling}.
We select DiT \cite{peebles2023scalablediffusionmodelstransformers}, SiT \cite{li2024scalable}, and MAR \cite{li2024autoregressiveimagegenerationvector} for downstream denoising-based image generation tasks.
For DiT and SiT, we set the patch size of DiT and SiT to 1 and use a 1D absolution position embedding.
In our main experiments, we train DiT-XL and SiT-XL of 675M parameters for 3M steps, compared to 4M steps in REPA \cite{yu2024representation} and 7M steps vanilla version \cite{peebles2023scalablediffusionmodelstransformers,ma2024sit}.
We strictly follow their original training setup for other settings.
For MAR-H, we train for 500 epochs and set the maximum learning rate to 2$e$-4 since a higher learning rate leads to NAN issues.  
For other experiments, we simply train SiT-L of 458M parameters for 400K steps.
More experimental details are provided in \cref{sec:appendix-exp-gen}.

\noindent \textbf{Evaluation}.
We use reconstruction Frechet Inception Distance (rFID) \cite{heusel2017gans} on ImageNet validation set to evaluate the tokenizer.
To evaluate the performance of generation tasks, we report generation FID (gFID), Inception Score (IS) \cite{salimans2016improved}, Precision and Recall \cite{kynkaanniemi2019improved} (in \cref{sec:appendix-results-gen}), with and without classifier-free guidance (CFG) \cite{ho2022classifier}. 
We measure the efficiency of the generative models by GLOPs of the model's forward pass on the latent codes of tokenizers, and training and inference throughput for the models using floating point 32 and a batch size of $64$ on a single AMD MI250.

\subsection{Main Results}
\label{sec:exp-main}

We present the main reconstruction and generation results on the ImageNet benchmark of resolution 256$\times$256 and 512$\times$512 in \cref{tab:main_256} and \cref{tab:main_512}, respectively.
We show that \softvq achieves reconstruction and generation performance (with generative models) comparable to leading systems with 32 and 64 tokens, while presenting a significant improvement in both training and inference efficiency.

\noindent \textbf{Consistent high-quality reconstruction}.
SoftVQ variants achieve remarkable reconstruction quality with a high compression ratio.
With only 32 and 64 tokens, our models maintain consistent rFID scores, \ie, 0.61-0.89 on the 256$\times$256 and 0.64-0.71 on 512$\times$512 benchmark, substantially outperforming VQ tokenizers with 256 tokens used in autoregressive models such as MaskGIT \cite{chang2022maskgitmaskedgenerativeimage} and TiTok \cite{yu2024an}.
Our results are also comparable to KL tokenizers used in previous denoising-based models, using \textbf{32x} fewer tokens.  
More discussion of different tokenizers is given in \cref{sec:exp-compare}.

\begin{table*}[t!]
\centering
\caption{\textbf{System-level comparison} on ImageNet 512$\times$512 conditional generation. We compare with both diffusion-based models and auto-regressive models with different types of tokenizers.  \colorbox{lightgray!20}{Grey} denotes SoftVQ. $^\dagger$ indicates results with DPM-Solver and 50 steps. }
\vspace{-0.1in}
\label{tab:main_512}
\resizebox{0.75\linewidth}{!}{%
\begin{tabular}{@{}lccc|cc|cc|cc@{}}
\toprule
\multirow{2}{*}{Gen. Model} &
  \multirow{2}{*}{Tok. Model} &
  \multirow{2}{*}{\# Tokens $\downarrow$} &
  \multirow{2}{*}{Tok. rFID  $\downarrow$} &
  \multirow{2}{*}{Gflops $\downarrow$} &
  \multirow{2}{*}{\begin{tabular}[c]{@{}c@{}}Throughput\\ (imgs/sec)\end{tabular}  $\uparrow$} &
  \multicolumn{2}{c|}{w/o CFG} &
  \multicolumn{2}{c}{w/ CFG} \\ 
 &
   &
   &
   &
   &
   &
  gFID  $\downarrow$ &
  IS  $\uparrow$ &
  gFID  $\downarrow$ &
  IS $\uparrow$ \\ \toprule

\multicolumn{10}{c}{\textbf{Generative Adversarial Models}} \\ \bottomrule

BigGAN \cite{chang2022maskgitmaskedgenerativeimage} &
  - &
  - &
  - &
  - &
  - &
  - &
  - &
  8.43 &
  177.9 \\

StyleGAN-XL \cite{karras2019style} &
  - &
  - &
  - &
  - &
  - &
  - &
  - &
  2.41 &
  267.7 \\ \bottomrule

\multicolumn{10}{c}{\textbf{Auto-regressive Models}} \\ \toprule

MaskGIT \cite{chang2022maskgitmaskedgenerativeimage} &
  VQ &
  1024 &
  1.97 &
  - &
  - &
  7.32 &
  156.0 &
  - &
  - \\
TiTok-B \cite{yu2024an} &
  VQ &
  128 &
  1.52 &
  - &
  - &
  - &
  - &
  2.13 &
  261.2 \\  \midrule 
 \cellc MAR-H &
\cellc  SoftVQ-BL &
  \cellc 64 &
\cellc  0.71 &
\cellc  86.55 &
\cellc  1.50 &
\cellc  8.21 &
\cellc  152.9 &
\cellc  3.42 &
\cellc  261.8 \\ \toprule
\multicolumn{10}{c}{\textbf{Diffusion-based Models}} \\ \bottomrule
ADM \cite{dhariwal2021diffusionmodelsbeatgans} &
  - &
  - &
  - &
  - &
  - &
  23.24 &
  58.06 &
  3.85 &
  221.7 \\
U-ViT-H/4$^\dagger$ \cite{bao2023all} &
  \multicolumn{1}{c}{\multirow{3}{*}{KL}} &
  \multirow{3}{*}{4096} &
  \multicolumn{1}{c|}{\multirow{3}{*}{0.62}} &
  128.92 &
  0.58 &
  - &
  - &
  4.05 &
  263.8 \\

DiT-XL/2 \cite{peebles2023scalablediffusionmodelstransformers} &
   &
   &
   &
  373.34 &
  0.10 &
  9.62 &
  121.5 &
  3.04 &
  240.8 \\

SiT-XL/2 \cite{ma2024sit} &
   &
   &
   &
  373.32 &
  0.10 &
  - &
  - &
  2.62 &
  252.2 \\  

DiT-XL \cite{chen2024deep} &
  \multicolumn{1}{c}{\multirow{2}{*}{AE}} &
  \multicolumn{1}{c}{\multirow{2}{*}{256}} &
  \multicolumn{1}{c|}{\multirow{2}{*}{0.22}} &
  80.75 &
  1.02 &
  9.56 &
  - &
  2.84 &
  - \\

UViT-H$^\dagger$ \cite{chen2024deep} &
  &
   &
   &
  128.90 &
  6.14 &
  9.83 &
  - &
  2.53 &
  - \\

UViT-H$^\dagger$ \cite{chen2024deep} &
  \multicolumn{1}{c}{\multirow{2}{*}{AE}} &
  \multicolumn{1}{c}{\multirow{2}{*}{64}} &
  \multicolumn{1}{c|}{\multirow{2}{*}{0.22}}  &
  32.99 &
  15.23 &
  12.26 &
  - &
  2.66 &
  - \\

UViT-2B$^\dagger$ \cite{chen2024deep} &
  &
   &
   &
  104.18 &
  9.44 &
  6.50 &
  - &
  2.25 &
  - \\

  \midrule
\cellc SiT-XL &
\cellc  SoftVQ-BL &
\cellc  64 &
\cellc  0.71 &
\cellc  29.23 &
\cellc  5.38 &
\cellc  7.96 &
\cellc  133.9 &
\cellc  2.21 &
\cellc  290.5 \\
 \bottomrule
\end{tabular}%
}
\vspace{-0.2in}
\end{table*}

\noindent \textbf{Generation performance comparable to SOTA}.
Notably, with only 32 tokens, the generation performance of DiT-XL and SiT-XL trained on all variants of the proposed SoftVQ significantly outperforms DiT-XL/2 and SiT-XL/2 with KL tokenizers of 1024 tokens, without using CFG. 
DiT-XL with SoftVQ-L of 32 tokens presents a \textbf{0.62} improvement on FID and SiT-XL with SoftVQ-L of 32 tokens shows a \textbf{0.71} improvement on FID,  without using CFG. 
The improvement is further strengthened with 64 tokens.
SiT-XL with SoftVQ-L of 64 tokens outperforms SiT-XL/2 with REPA \cite{yu2024representation}, without using CFG.
Applying CFG, SiT-XL with SoftVQ of 64 tokens achieves a performance comparable to the leading systems with the best FID as \textbf{1.78} on 256 and \textbf{2.21} on 512 benchmark.
Using MAR-H \cite{li2024autoregressiveimagegenerationvector}, SoftVQ-BL and SoftVQ-L with 64 and 32 tokens present slightly worse results, compared to the KL tokenizer of 256 tokens, possibly due to the lowered learning rate.
MAR-H and SiT-XL results on 512 benchmark with 32 tokens are included in \cref{sec:appendix-results-main}.

\noindent \textbf{Efficiency}.
The system-level comparison reveals SoftVQ's exceptional efficiency across multiple metrics. 
When integrated with MAR-H at 256$\times$256 resolution, SoftVQ reduces GFLOPs by ~\textbf{40}\% compared to standard KL tokenizer, \ie, 86.55 vs 145.08 GFLOPs with 64 tokens and further reducing to 67.93 GFLOPs with 32 tokens. 
SoftVQ enables DiT-XL and SiT-XL to achieve significantly lower computational costs \ie, 28.81 and 29.23 GFLOPs, respectively with 64 tokens, compared to their KL counterparts (80.73 GFLOPs with 1024 tokens).
Importantly, these efficiency improvements come without compromising generation quality.
SiT-XL trained on SoftVQ-BL with 64 tokens presents ~\textbf{55x} throughput than the KL counterpart, while showing competitive IS and FID scores in all configurations.
Not only do we require fewer training iterations to achieve a comparable performance to leading systems, \ie, 2.5M vs 7M steps of SiT-XL and DiT-XL, the high compression ratio of SoftVQ also improves the training throughput, reducing the time to train SiT-XL for 400K steps from 72 to 20 hours on 8 MI250.


\subsection{Comparison of Tokenizers}
\label{sec:exp-compare}

We compare the proposed \softvq with the concurrent efficient image tokenizers, \ie, TiTok \cite{yu2024an} and DC-AE \cite{chen2024deep}. 
To show the superiority of \softvq in achieving a high compression ratio, we additionally train \vq and AE using the small (S) configuration and the same training recipe of SoftVQ. 
To validate the generation performance, we train a SiT-L for 400K steps and report gFID and IS on 256 ImageNet without using CFG, as shown in \cref{tab:tokenizer_compare}. 

\noindent \textbf{Scalable performance}.
With a much smaller model size, \ie, 46M vs 390M, SoftVQ-S significantly outperforms both TiTok variants at 64 tokens, achieving better reconstruction quality with an rFID of \textbf{1.03} compared to 1.25, and better generation performance with a gFID of \textbf{11.24} and IS of \textbf{89.4} compared to the best TiTok gFID of 19.23 and IS of 61.8. 
Noteworthy is that the generation results SiT-L trained on SoftVQ with 32 tokens outperform those trained on KL tokenizer with 1024 tokens \cite{ma2024sit} by a large margin, \ie, \textbf{5.9} in gFID.
Compared to DC-AE, SoftVQ-S demonstrates competitive scalability across token counts while maintaining significant generation quality even with fewer parameters.

\begin{table}[t!]
\centering
\caption{Comparison of tokenizers on class-conditional 256$\times$256 ImageNet. We report rFID, and gFID and IS of SiT-L without CFG.}
\label{tab:tokenizer_compare}
\vspace{-0.1in}
\resizebox{0.98\columnwidth}{!}{%
\begin{tabular}{@{}l|ccc|c|cc@{}}
\toprule
\multirow{2}{*}{Tokenizer} & \multirow{2}{*}{\# Params} & \multirow{2}{*}{\# Tokens} & \multirow{2}{*}{Dim.} & \multirow{2}{*}{rFID $\downarrow$} & \multicolumn{2}{c}{SiT-L} \\
                          &                      &      &     &       & gFID $\downarrow$  & IS $\uparrow$  \\ \midrule
KL  \cite{stabilityai2023}                      & 676M                 & 1024 & 4   & 0.62  & 18.79 & 72.0 \\ \midrule
TiTok-BL-KL \cite{yu2024an}             & 389M                 & 64   & 16  & 1.25  &   23.35    &  54.7    \\
TiTok-BL-VQ  \cite{yu2024an}              & 390M                 & 64   & 16  & 2.06  &   19.23    &   61.8   \\ \midrule
DC-AE-f32  \cite{chen2024deep}               & 323M                 & 64   & 32  & 0.69  &  15.36     &  76.1    \\
DC-AE-f54  \cite{chen2024deep}                & 677M                 & 16   & 128 & 0.81  &    24.66   &   49.1   \\ \midrule
\multirow{4}{*}{VQ-S}     & \multirow{4}{*}{46M} & 256  & 8   & 1.45  &   19.02    &  58.6    \\
                          &                      & 128  & 16  & 2.61  &   18.51    &  62.4    \\
                          &                      & 64   & 32  & 4.04  &   20.07    &  54.5    \\
                          &                      & 32   & 64  & 10.97 &   21.57    &  47.6    \\ \midrule
\multirow{4}{*}{AE-S}     & \multirow{4}{*}{46M} & 256  & 8   & 1.15  &    22.11   &   58.3   \\
                          &                      & 128  & 16  & 1.33  &    22.31   &    58.5  \\
                          &                      & 64   & 32  & 1.64  &     25.53  &   54.3  \\
                          &                      & 32   & 64  & 2.01  &     28.18  & 45.5    \\ \midrule
\multirow{4}{*}{SoftVQ-S} & \multirow{4}{*}{46M} & 256  & 8   & 0.80  &  9.21     &   93.6   \\
                          &                      & 128  & 16  & 0.92  &   10.12    &   85.8   \\
                          &                      & 64   & 32  & 1.03  &  11.24     &  89.4    \\
                          &                      & 32   & 64  & 1.24  &    12.89   &   79.5   \\ \bottomrule
\end{tabular}%
}
\vspace{-0.25in}
\end{table}

\noindent \textbf{Less Lossy Property}.
SoftVQ-S exhibits remarkably consistent performance across different compression ratios. 
When reducing tokens from 256 to 32, SoftVQ-S maintains a minimal degradation in rFID from 0.80 to 1.24, significantly outperforming both VQ-S, from 1.45 to 10.97, and AE-S, from 1.15 to 2.01.
The benefits of the fully-differentiable property of SoftVQ along with representation alignment are particularly evident in generation quality, where SoftVQ-S achieves substantially better gFID of 9.21-12.89, and IS scores of 79.5-93.6, compared to baselines, showing its robustness in preserving information at high compression ratios and a better latent space for generative models.

\subsection{Discussions on the Latent Space}
\label{sec:exp-latent}

We perform more analysis on the latent of SoftVQ here.

\noindent \textbf{Alignment with different initialization and target models}.
As shown in \cref{tab:represent_align}, our results demonstrate the effectiveness of alignment across various pre-trained models. 
The encoder initialization and alignment with DINOv2-B \cite{oquab2023dinov2} achieve superior performance with rFID 0.88 and IS 103.4, compared to using either component alone.
Further improvements are observed with CLIP-B \cite{radford2021learning} and EVA-02-B \cite{fang2024eva} on reconstruction.
We reveal that a better rFID does not necessarily translate to a better gFID.
Instead, the latent space quality, reflected by linear probing accuracy, is more closely related to the performance of the generative models.

\begin{figure}[t!]
    \centering
    \includegraphics[width=0.9\linewidth]{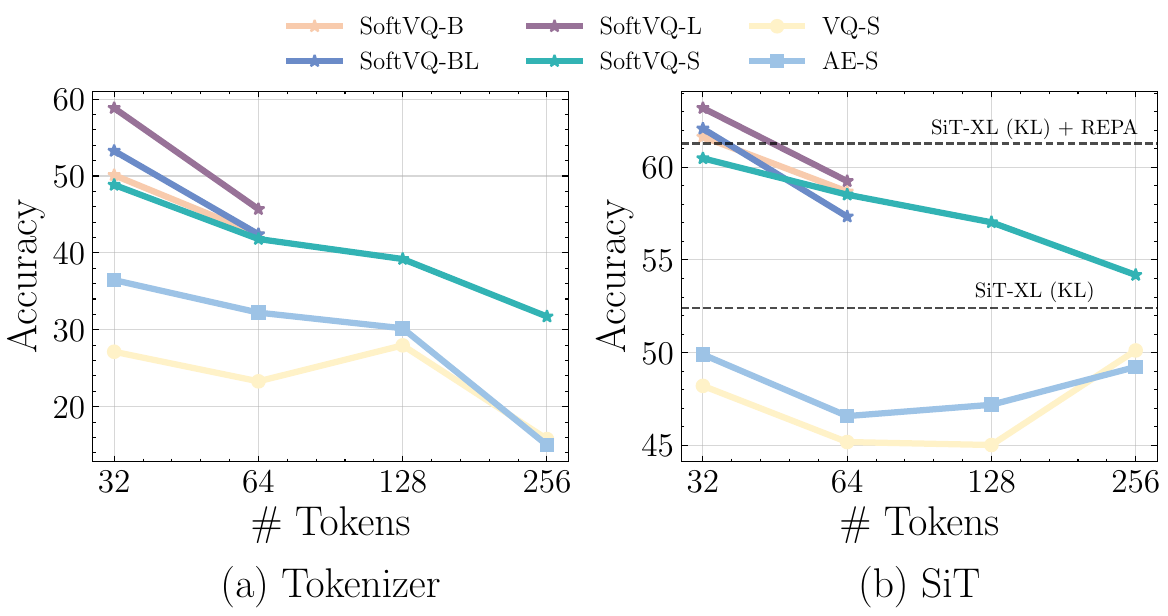}
    \vspace{-0.15in}
    \caption{Linear probing accuracy of ImageNet-1K val. set on (a) latent tokens of tokenizer and (b) intermediate features (layer 20) of SiT (L for small and XL for others) trained on latents of tokenizer.}
    \label{fig:linear_probing}
\vspace{-0.15in}
\end{figure}

\noindent \textbf{Linear probing of tokenizer and generative model}.
Linear probing results in \cref{fig:linear_probing} (details in \cref{sec:appendix-exp-lp}) reveal SoftVQ's superior representation quality at both tokenizer and generative model. 
At the tokenizer level, the linear probing accuracy of SoftVQ variants (B, BL, L) consistently outperforms VQ-S and AE-S across all token counts, with the gaps widening at lower token counts, similarly observed by \citet{yu2024an}. 
This advantage transfers to the generative model with the best linear probing accuracy on the intermediate features of SiT trained with SoftVQ.

\begin{figure}[t]
    \centering
    \includegraphics[width=0.9\linewidth]{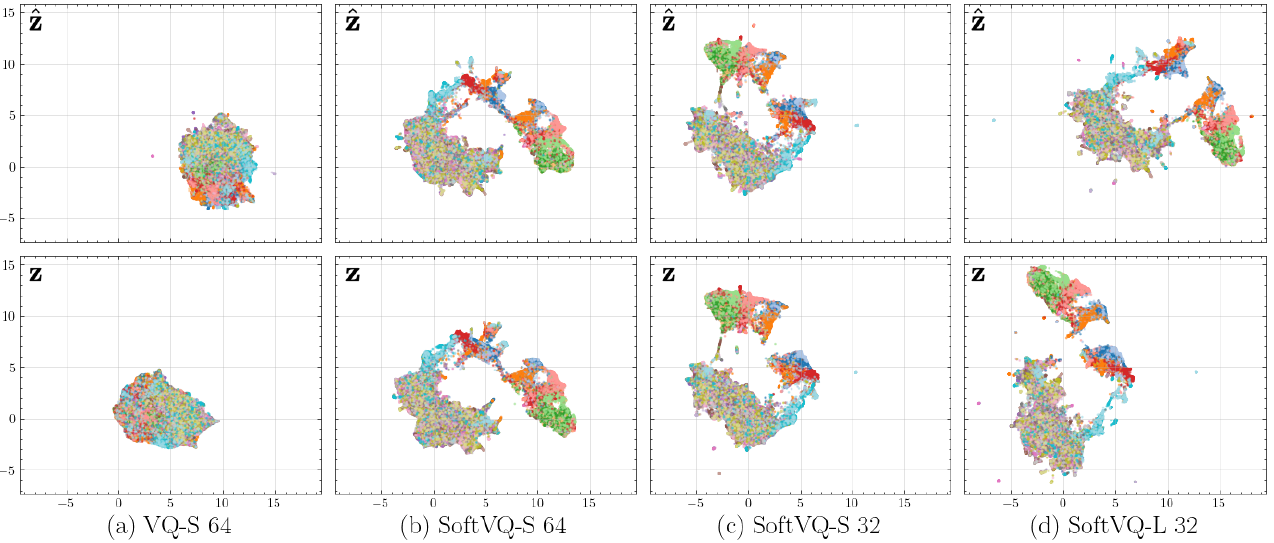}
    \vspace{-0.1in}
    \caption{Visualization of $\hat{\mathbf{z}}$ (top), \ie, encoder output, and $\mathbf{z}$ (bottom), \ie, decoder input, of (a) VQ-S 64; (b) SoftVQ-S 64; (c) SoftVQ-S 32; (d) SoftVQ-L 32, trained with latent space alignment.
    }
    \label{fig:latent_vis}
\vspace{-0.2in}
\end{figure}

\noindent \textbf{Latent space visualization}.
UMAP visualizations on the latent space in \cref{fig:latent_vis} demonstrate SoftVQ's ability to maintain structured and discriminative latent representations with minimal variance between encoder output ($\hat{\mathbf{z}}$) and decoder input ($\mathbf{z}$). 
Comparing \cref{fig:latent_vis} (d) and (c), we also observe the larger encoder can learn a more discriminative latent space, showing the superiority of SoftVQ's representation learning.


\subsection{Ablation Studies}
\label{sec:exp-ablation}

We present a series of ablation studies in \cref{sec:appendix-results-ablation}, including SoftVQ variants with product quantization (PQ) \cite{jegou2010product}, residual quantization (RQ) \cite{lee2022autoregressive}, and GMMVQ, codebook size, latent space size, and the temperature of SoftVQ.
The results show the compatibility of SoftVQ with PQ and RQ, with slightly improved performance.
GMMVQ presents results comparable to SoftVQ.
Moreover, SoftVQ shows robustness to other parameters, such as codebook size and softmax temperature. 
We found that while increasing the dimension of the latent space can result in a significant improvement in reconstruction, it leads to deterioration in generation performance, possibly due to the learning difficulty of generative models with larger dimensions \cite{sohl2015deep}.

\begin{table}[t!]
\centering
\caption{Representation alignment with different encoder initialization and the target alignment models using SoftVQ-B 64 tokens.}
\vspace{-0.1in}
\label{tab:represent_align}
\resizebox{0.8 \columnwidth}{!}{%
\begin{tabular}{@{}cc|cc|cc@{}}
\toprule
\multirow{2}{*}{Enc. Init.} & \multirow{2}{*}{Align. Target} & \multirow{2}{*}{rFID $\downarrow$} & \multirow{2}{*}{L.P. $\uparrow$} & \multicolumn{2}{c}{SiT-L} \\
         &          &  &  & gFID $\downarrow$ & IS  $\uparrow$\\ \midrule
-        & -        & 0.98 & 5.42  & 20.33     & 89.4   \\
-        & DINOv2-B \cite{oquab2023dinov2}   & 0.93 &  41.08 &  10.96   & 101.9   \\
DINOv2-B \cite{oquab2023dinov2}   & -       & 0.73 & 11.87  &  17.20    &   91.5 \\
DINOv2-B \cite{oquab2023dinov2}   & DINOv2-B \cite{oquab2023dinov2}   & 0.88 &  42.42 &   10.13   &  103.4  \\
CLIP-B \cite{radford2021learning}  & CLIP-B \cite{radford2021learning}   &  0.75 & 47.12 & 9.89  & 109.3     \\
EVA-02-B \cite{fang2024eva} & EVA-02-B \cite{fang2024eva} &  0.81 & 44.23 &   10.34   &  112.7  \\ \bottomrule
\end{tabular}%
}
\vspace{-0.25in}
\end{table}

\section{Related Work}
\label{sec:related}

\noindent \textbf{Image tokenization} has emerged as a fundamental technique to bridge various vision tasks. Traditional approaches using autoencoders \cite{hinton2006reducing,vincent2008extracting} laid the groundwork by compressing images into latent representations. This field subsequently diverged into two main branches: generation-focused and understanding-focused approaches. Generation-oriented methods, such as VAE \cite{van2017neural,razavi2019generating} and VQ-GAN \cite{esser2021taming,razavi2019generatingdiversehighfidelityimages}, emphasized learning latent spaces for detail-preserving compression. These were further refined through variants \cite{lee2022autoregressive,yu2023language,mentzer2023finite,zhu2024scaling} that enhanced generation quality. Parallel developments in understanding-focused approaches leveraged Large Language Models (LLMs) \cite{vaswani2023attentionneed} to create semantic representations for tasks like classification \cite{dosovitskiy2021imageworth16x16words} and detection \cite{zhu2010deformable}. Recent work \cite{yu2024an,li2024imagefolder} has demonstrated the potential of unifying these approaches, with subsequent research \cite{wu2024vila,gu2023rethinkingobjectivesvectorquantizedtokenizers} exploring the convergence of generation and understanding capabilities using a single tokenizer.

\noindent \textbf{Image generation} have two popular paradigms: autoregressive and diffusion. 
Autoregressive models start from CNN-based architectures \cite{van2016conditional} and evolve to the transformer-based architectures \cite{vaswani2023attentionneed,yu2024randomized,lee2022autoregressive,liu2024customize,sun2024autoregressive} to enhance scalability. Latest works in visual autoregressive modeling, including VAR \cite{tian2024visualautoregressivemodelingscalable}, MAR \cite{li2024autoregressiveimagegenerationvector}, and ImageFolder \cite{li2024imagefolder}, have further enhanced generation efficiency. 
Diffusion models have transformed image generation since their introduction by Sohl-Dickstein et al. \cite{sohldickstein2015deepunsupervisedlearningusing}. Key advancements by Nichol et al. \cite{nichol2021improveddenoisingdiffusionprobabilistic}, Dhariwal et al. \cite{dhariwal2021diffusionmodelsbeatgans}, and Song et al. \cite{song2022denoisingdiffusionimplicitmodels} have significantly improved model performance. A paradigm shift occurred with the adoption of latent space diffusion \cite{vahdat2021scorebasedgenerativemodelinglatent,rombach2022highresolutionimagesynthesislatent}, leveraging pre-trained image encoders as priors \cite{van2017neural,esser2021taming} to enhance generation efficiency. Architectural innovations, particularly the integration of transformers \cite{peebles2023scalablediffusionmodelstransformers}, have further expanded the capabilities of these models.
These advances have spawned numerous influential systems, from text-guided generation \cite{nichol2021glide,ding2021cogview} to high-fidelity synthesis \cite{gafni2022make,saharia2022photorealistic}. 
Commercial successes \cite{ramesh2021zero,rombach2022high,midjourney,sora} have also demonstrated the practical impact of diffusion models.

\section{Conclusion}
\label{sec:conclusion}
We presented \softvq, a continuous image tokenizer leveraging soft categorical posteriors to aggregate multiple codewords into each latent token. 
Our approach compresses images to just 32 or 64 tokens while maintaining high reconstruction quality and enabling state-of-the-art generation results with DiT, SiT, and MAR. The substantial improvements in both inference throughput (up to 55x with 64 tokens) and training efficiency, combined with the fully-differentiable design's ability to learn semantic-rich representations, demonstrate \softvq's potential as a foundation for efficient generative and vision-language models.

{
    \small
    \bibliographystyle{ieeenat_fullname}
    \bibliography{main}
}

\clearpage
\setcounter{page}{1}
\maketitlesupplementary

\appendix

\section{Posterior of \kl and \vq}

In this section, we provide the detailed derivation of the KL-divergence of \kl and \vq used in \cref{eq:kl} and \cref{eq:vq}, respectively.

\noindent \textbf{\kl} has the KL divergence of the posterior with a Gaussian prior:

\begin{equation}
\begin{aligned}
& \mathcal{L}_{\mathrm{kl}} \left(q_{\phi}(\mathbf{z}) \| p(\mathbf{z})\right). \\
& =\int q_{\phi}(\mathbf{z})\left(\log q_{\phi}(\mathbf{z})  - \log p(\mathbf{z}) \right) d \mathbf{z} \\
& =\frac{1}{2} \sum_{j=1}^D\left(1+\log \left(\left(\sigma_j\right)^2\right)-\left(\mu_j\right)^2-\left(\sigma_j\right)^2\right),
\end{aligned}
\end{equation}
\noindent where $D$ is the latent space dimension.

\noindent \textbf{\vq} assumes a uniform prior over the total $K$ codewords in the learnable codebook for the deterministic posterior, thus present a KL divergence as follows:
\begin{equation}
\begin{aligned}
& \mathcal{L}_{\mathrm{kl}} \left(q_{\phi}(\mathbf{z}) \| p(\mathbf{z})\right). \\
& =\int q_{\phi}(\mathbf{z})\left(\log q_{\phi}(\mathbf{z})  - \log p(\mathbf{z}) \right) d \mathbf{z} \\
&=-(-\log K) \\
&=\log K \\
\end{aligned}
\end{equation}

\subsection{Comparison to VQ}

We list the advantages of SoftVQ in \cref{tab:advantage} with empirical support, and will add in the revision.

\begin{table}[h!]
\centering
\vspace{-0.1in}
\caption{Advantages of SoftVQ over VQ.}
\vspace{-0.1in}
\label{tab:advantage}
\resizebox{0.95\columnwidth}{!}{%
\begin{tabular}{@{}l|ccl@{}}
\toprule
Advantages        & VQ       & SoftVQ     & Empirical Support \\ \midrule
Grad. Broken & yes      & no         &   
\begin{tabular}{l}
     No straight-through trick. SoftVQ-S-64 has a 1.41 rFID and 13.35 gFID.
\end{tabular} \\ 
Codebook Loss   & yes      & no         &  
\begin{tabular}{l}
     Not necessary. Adding codebook loss in SoftVQ-S-64  \\
    results in an 1.34 rFID of  of but a worse gFID of 15.15.
\end{tabular}
 \\
Commit Loss     & yes      & no         &
\begin{tabular}{l}
     Not necessary. Adding commit loss in SoftVQ-S-64  \\
    results in an rFID of 1.33 of but a worse gFID of 14.35.
\end{tabular}
 \\
Comp. Ratio     & low      & high       &      
\begin{tabular}{l}
     SoftVQ has much lower rFID$<$1.0 comapred to TiTok in Tab. 1.
\end{tabular} 
\\
Rep. Align.     &   difficult       &   easy         &  
\begin{tabular}{l}
    More discriminative features in main Fig. 4  
\end{tabular} 
    \\ \bottomrule
\end{tabular}%
}
\vspace{-0.1in}
\end{table}

While representation alignment can also applied to VQ, it fails to learn discriminative features with high compression ratio, mainly due to its broken gradient. 
Compared to SoftVQ, whose encoder and codebook's parameters are directly learned by the alignment loss, the gradient of alignment loss on VQ will first be straight-through from decoder input to encoder output, and then the codebook is updated according to the codebook loss, resulting in indirect learning of discriminative features. 

\begin{figure}[b!]
    \centering
    \vspace{-0.25in}
    \includegraphics[width=0.8\linewidth]{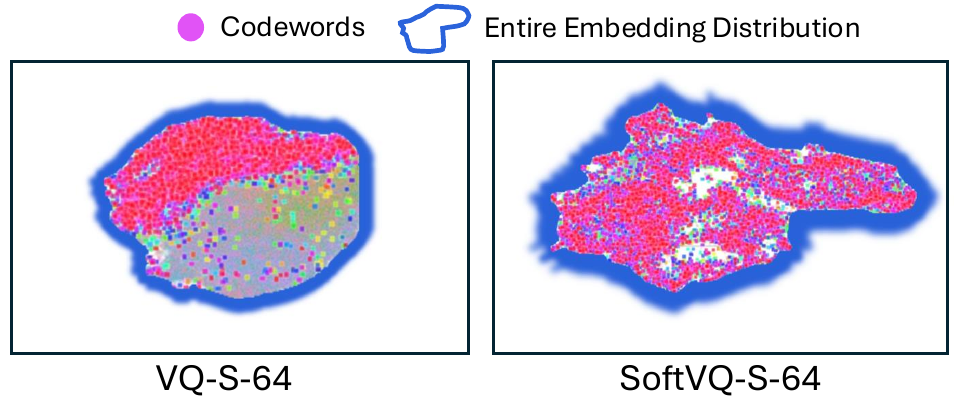}
    \vspace{-0.1in}
    \caption{Codewords visualization}
    \label{fig:codewords}
\end{figure}

We provide an additional codewords visualization in \cref{fig:codewords}, where SoftVQ learns code embeddings uniformly across the entire distribution. 

\section{Additional Details of SoftVQ-VAE and its Variants}
\label{sec:appendix-softvq}

In this section, we present more details on the posterior of \softvq, its variant GMMVQ-VAE with the latent space as a GMM model, and the compatibility of \softvq with improvement techniques of VQVAE.

\subsection{Posterior of \softvq}
\label{sec:appendix-softvq-kl}

In SoftVQ, we similarly assume that the prior is a uniform distribution over the $K$ learnable codewords as in VQ, except for we have the posterior as the softmax probability:

\begin{equation}
\begin{aligned}
    & \mathcal{L}_{\mathrm{kl}} \left(q_{\phi}(\mathbf{z}) \| p(\mathbf{z})\right). \\
    & =\int q_{\phi}(\mathbf{z})\left(\log q_{\phi}(\mathbf{z})  - \log p(\mathbf{z}) \right) d \mathbf{z} \\
    &= H(q_{\phi}(\mathbf{z})) - H(  q_{\phi}(\mathbf{z}), p(\mathbf{z})),
\end{aligned}
\end{equation}
where $H(q_{\phi}(\mathbf{z})$ is the entropy for $H( q_{\phi}(\mathbf{z}), p(\mathbf{z}))$ is the cross-entropy between $q_{\phi}(\mathbf{z})$ and the uniform prior $p(\mathbf{z}) \sim \mathcal{U}(0, K)$.
In practice, $H( \mathbb{E}_{\mathbf{x} \sim p(\mathbf{x})} [ q_{\phi}(\mathbf{z})], p(\mathbf{z}))$ we compute the $\mathbb{E}_{\mathbf{x} \sim p(\mathbf{x})} [ q_{\phi}(\mathbf{z})]$ instead, where $\mathbb{E}_{\mathbf{x} \sim p(\mathbf{x})} [ q_{\phi}(\mathbf{z})]$ is computed on the averaged batch data during training.

\subsection{GMMVQ-VAE}
\label{sec:appendix-softvq-gmm}

As discussed in the main paper, the latent space of SoftVQ can be interpreted as soft K-Means, and it can be further extended to a latent space of Gaussian Mixture Model. We present more details of this extension here.

In GMMVQ-VAE, the encoder predicts two things: the embedding $\hat{\mathbf{z}}$ and the weights of the Gaussian component $\omega(\hat{\mathbf{z}})$. We can then formulate the posterior as:
\begin{equation}
\begin{split}
    & \textit{posterior: }  q_{\phi}(\mathbf{z} | \mathbf{x}) = \softmax \left(- \omega(\hat{\mathbf{z}}) \| \hat{\mathbf{z}} - \mathcal{C} \|_2\right) \\
    & \textit{latent: } \mathbf{z} = q_{\phi}(\mathbf{z} | \mathbf{x}) \mathcal{C} \\ 
    & \textit{kl: } \mathcal{L}_{\mathrm{kl}} = H(q_{\phi}(\mathbf{z} | \mathbf{x}))- H\left(\mathbb{E}_{\mathbf{x} \sim p(\mathbf{x})} q_{\phi}(\mathbf{z} | \mathbf{x})\right). \\ 
\end{split}
\end{equation}
The difference between SoftVQ and GMMVQ in our formulation is the way in computing the posterior, \ie, using fixed temperature parameter versus learning the data-dependent temperature parameters by the encoder. 
Note that we still maintain the codebook and its codewords entries directly for the decoder input. 
It is possible to make the latent space formally a GMM, by treating the codewords as Gaussian means, and formulating the decoder input with the re-parametrization trick. 
However, we find that this formulation with re-parametrization will hinder the learning of the latent space.
Thus, we adopted the simpler design to use the codebook directly for reconstruction, assuming fixed variance in the Gaussian components of GMM.

\subsection{Compatibility of \softvq}
\label{sec:appendix-softvq-compat}

\begin{figure*}
    \centering
    \includegraphics[width=0.85\linewidth]{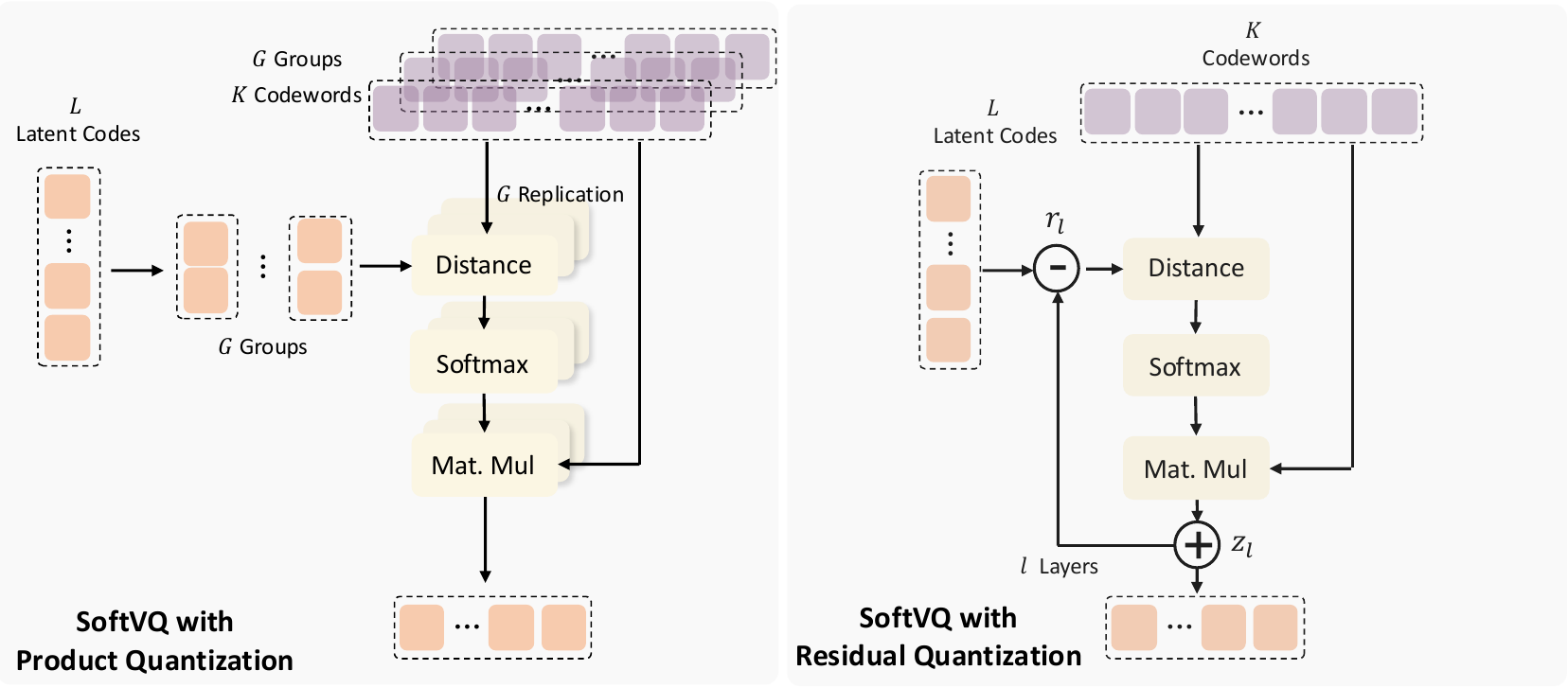}
    \caption{Illustration of product quantization (left) and residual quantization (right) with SoftVQ.}
    \label{fig:appendix-prod-res}
\end{figure*}

Since \softvq maintains the learnable codebook, previous techniques to improve \vq are also compatible with it. 
Here, we follow ImageFolder \cite{li2024imagefolder} to show the combination of SoftVQ with product quantization \cite{jegou2010product} and residual quantization \cite{lee2022autoregressive}, as illustrated in \cref{fig:appendix-prod-res}.

\noindent \textbf{Product Quantization (PQ)} \cite{jegou2010product} divides the latent codes into $G$ groups, with each group having its own codebook:
\begin{equation}
\begin{split}
    \mathbf{z} &= [\mathbf{z}^{(1)}, \mathbf{z}^{(2)}, ..., \mathbf{z}^{(G)}] \\
    \quad q_{\phi} (\mathbf{z}^{(g)}|\mathbf{x}) &= \softmax(- \|\hat{\mathbf{z}}^{(g)} - \mathcal{C}^{(g)} \|_2 / \tau ).
\end{split}
\end{equation}
PQ can effectively increase the actual codebook size.

\noindent \textbf{Residual Quantization (RQ)} \cite{lee2022autoregressive} applies multiple layers of quantization to the residual errors of the encoder output:
\begin{equation}
\begin{split}
\mathbf{z}_l &= \mathbf{z}_{l-1} + \operatorname{SoftVQ}(\mathbf{r}_{l-1})  \\
 \mathbf{r}_l &= \mathbf{r}_{l-1} - \mathbf{z}_l \\
\mathbf{z}_0 & = 0 \\  
\mathbf{r}_0 &= \hat{\mathbf{z}}
\end{split}
\end{equation}
where $\mathbf{r}$ is the residual and $l$ is the layer index.
RQ captures more fine-grained features and thus better reconstruction.

\section{Additional Implementation Details}
\label{sec:appendix-exp}

In this section, we provide more implementation details of the tokenizer training, the downstream generative models training, and the linear probing evaluation.

\subsection{Implementation Details of \softvq}
\label{sec:appendix-exp-softvq}

All tokenizer experiments (except the open-source ones) in this paper are trained using the same training recipe. 
We train the tokenizers on ImageNet \cite{deng2009imagenet} of resolution 256$\times$256 and 512$\times$512 for 250K iterations on 8 MI250 GPUs.
Training for longer may lead to further improvement of reconstruction and potentially downstream generation performance \cite{luo2024open,weber2024maskbit}.
AdamW \cite{loshchilov2017decoupled} optimizer is used with $\beta_1=0.9$, $\beta_2=0.95$, a weight decay of 1$e$-4, a maximum learning rate of 1$e$-4, and cosine annealing scheduler with a linear warmup of 5K steps.
We use a constant batch size of 256 is used for all models.
For the training objective, we set $\lambda_1=1.0$, $\lambda_1=0.2$, $\lambda_3=0.1$, and $\lambda_4=0.01$, following previous common practice. 
We additionally linearly warmup the loss weight of perceptual loss, \ie, $\lambda_1=1.0$, for 10K iterations, which we find beneficial to train with fewer latent tokens.

Similarly to \citet{tian2024visualautoregressivemodelingscalable} and \citet{li2024imagefolder}, we found that the discriminator is very important for training the tokenizer.
Instead of using the common PatchGAN and StyleGAN architecture, we adopted the frozen DINO-S  \cite{caron2021emerging,oquab2023dinov2} discriminator with a similar architecture to StyleGAN \cite{karras2019style,karras2020analyzing}, as in \cite{tian2024visualautoregressivemodelingscalable}.
In addition, we use DiffAug \cite{zhao2020differentiable}, consistency regularization \cite{zhang2019consistency}, and LeCAM regularization \cite{tseng2021regularizinggenerativeadversarialnetworks} for discriminator training as in \cite{tian2024visualautoregressivemodelingscalable}.
The loss weight of consistency regularization is set to 4.0 and LeCAM regularization is set to 0.001.
We did not use the adaptive discriminator loss weight since it is too tricky to tune.

\subsection{Implementation Details of DiT, SiT, and MAR}
\label{sec:appendix-exp-gen}

\noindent \textbf{DiT}. 
The training recipe of our DiT models mainly follows the original setup. 
Since we are using 1D latent tokens, we set the patch size of DiT models to 1 and use 1D absolute position encoding. 
We use a constant learning rate of 1$e$-4 and a global batch size of 256 to train the DiT models.
We use the cosine noise scheduler since it suits better for our case, but very similar generation performance were obtained in our early experiments with linear scheduler.
To accelerate training, we additionally adopt mixed precision with bfloat16 and flash-attention.
DiT-L models are trained for 400K iterations. 
In our main paper, we report the results of DiT-XL models for training 3M iterations. 
We report more and better results for longer training, \ie, up to 4M iterations, in \cref{sec:appendix-results-main}.
For conditional generation with CFG, we use 1.35 for DiT models trained on SoftVQ with 64 tokens and 1.45 for DiT models trained on SoftVQ with 32 tokens.
These guidance scale values are obtained via grid search.

\noindent \textbf{SiT}.
Similarly, we use the original setup to train the SiT models, with a constant learning rate of 1$e$-4 and a global batch size of 256.
We also use cosine scheduler in training of SiT models, as in our DiT models.
The main results are reported with training for 3M iterations, and we include the training results for up to 4M iterations in \cref{sec:appendix-results-main}.
For conditional generation with CFG, we use 1.75 for SiT models trained on SoftVQ with 64 tokens and 2.25 for models trained on SoftVQ with 32 tokens.
Similarly to RPEA \cite{yu2024representation}, we set the guidance interval to [0, 0.7] for CFG results \cite{kynkaanniemi2024applying}.
These guidance scale values are obtained via grid search.

\noindent \textbf{MAR}. 
The MAR-H are trained using the AdamW optimizer for 500 epochs.
The weight decay and momenta for AdamW are 0.02 and (0.9, 0.95). 
We also use a batch size of 2048 as in the original setup.
\citet{li2024autoregressiveimagegenerationvector} used a linearly scaled learning rate of 8$e$-4. 
However, we found that this learning rate constantly causes the NaN problems in loss.
Thus we adopt a maximum learning rate of 2$e$-4.
We also train the models with a 100-epoch linear warmup of the learning rate, followed by a constant schedule.
We suspect that the lower performance of our MAR models is caused by the lower learning rate, which effectively results in fewer training iterations. 
More investigation on improving the MAR performance of SoftVQ is undergoing and will be updated in our future iterations.
We use a CFG of 3.0 for conditional generation results.

We train all of our 256$\times$256 generative models on 8 MI250 GPUs and 512$\times$512 models on 8 MI300 GPUs.

\subsection{Implementation Details of Linear Probing}
\label{sec:appendix-exp-lp}

We use the setup similar to that used in MAE \cite{he2022masked}, DAE \cite{chen2024deconstructing}, and REPA \cite{baevski2020wav2vec} to perform the linear probing evaluation in the latent tokenizer space and intermediate features of trained SiT models. 
Specifically, we train a parameter-free batch normalization layer and a linear layer for 90 epochs with a batch size of 16,384.
We use the Lamb optimizer with cosine decay learning rate scheduler where the initial learning rate is set to 0.003.

\section{Additional Results}
\label{sec:appendix-results}

In this section, we present our ablation study on the SoftVQ-VAE tokenizer, more results of the trained generative models including results from longer training, and more visualization of the reconstruction results from the tokenizer and the generative models.

\subsection{Ablation Study}
\label{sec:appendix-results-ablation}

\begin{table}[t!]
\centering
\caption{Ablation study of SoftVQ-VAE}
\vspace{-0.1in}
\label{tab:ablation}
\resizebox{\columnwidth}{!}{%
\begin{tabular}{@{}ccccccc@{}}
\toprule
\multicolumn{7}{c}{Variants} \\ 
\bottomrule
\multicolumn{1}{c|}{Tokenizer} & SoftVQ-S                  & + PQ (G=2) & + PQ (G=4) & + RQ & + R-PQ (G=4) & GMMVQ-S \\ \midrule
\multicolumn{1}{c|}{rFID}      &  1.33 &    1.19        &    1.03                   &    1.55  &    1.17          &   1.12     \\  \bottomrule
\multicolumn{7}{c}{Codebook Size} \\ \toprule
\multicolumn{1}{c|}{$K$} & 512   & 1024 & 2048 & 4096 & 8192 & 16384 \\ \midrule
\multicolumn{1}{c|}{rFID}      & 1.82 &  1.58          &   1.35                    &  1.14    &  1.03       &   1.23     \\  \bottomrule
\multicolumn{7}{c}{Latent Size} \\ \toprule
\multicolumn{1}{c|}{$L$/$D$} & 64/16   & 64/32 & 64/64 & 32/32 & 32/64 & 32/128 \\ \midrule
\multicolumn{1}{c|}{rFID}      & 2.31 &      1.03     &     0.63                  &  1.61    &    1.24          &    0.79    \\  \bottomrule
\multicolumn{7}{c}{Softmax Temp.} \\ \toprule
\multicolumn{1}{c|}{$\tau$} & 0.001   & 0.01 & 0.07 & 0.1 & 1.0 & learnable \\ \midrule
\multicolumn{1}{c|}{rFID}      & 2.03 &  1.37           &   1.03                    &   1.19   &    1.52          &  1.10      \\  \bottomrule

\end{tabular}%
}
\vspace{-0.2in}
\end{table}

The full ablation results are shown in \cref{tab:ablation}.

\noindent \textbf{SoftVQ variants}. We compare our baseline SoftVQ-S with several variants, including different quantization approaches. Adding product quantization (PQ) with group size G=2 and G=4 both improve the performance, reducing rFID from 1.33 to 1.19 and 1.03 respectively. Although residual quantization (RQ) alone increases rFID to 1.55, combining it with product quantization (R-PQ) yields better results with rFID of 1.17. The GMMVQ-S variant also shows competitive performance with an rFID of 1.12.

\noindent \textbf{Codebook size}. We investigate the impact of codebook size K ranging from 512 to 16384. The results show that larger codebooks generally lead to better performance, with rFID decreasing from 1.82 (K=512) to 1.03 (K=8192). However, further increasing K to 16384 slightly degrades performance with an rFID of 1.23, suggesting that an overly large codebook may harm model stability.

\noindent \textbf{Latent size}. The latent dimensions L/D significantly affect the model performance. Starting from a small latent size of 64/16, increasing the dimensions initially helps, with 64/32 achieving the best rFID of 1.03. 
While further enlargement leads to better reconstruction performance, as seen in the 64/64 configuration with an rFID of 0.63, we find that the larger dimension of the latent space makes the generative models more challenging to learn.

\noindent \textbf{Softmax temperature}.
The softmax temperature $\tau$ controls the sharpness of the posterior probability. 
We observe that very low temperatures $\tau=$0.001 result in poor performance, while moderate values around 0.07 achieve optimal results. 
Making $\tau$ learnable produces competitive performance, suggesting that adaptive temperature could be a practical choice.

\subsection{Main Results}
\label{sec:appendix-results-main}

\begin{table*}[t!]
\centering
\caption{\textbf{System-level comparison} on ImageNet 256$\times$256 conditional generation. We compare with both diffusion-based models and auto-regressive models with different types of tokenizers. \colorbox{lightgray!20}{Grey} denotes SoftVQ. 
Our DiT/SiT results are reported with a total training of 3M iterations (compared to 4M of SiT/XL-2 + REPA and 7M for SiT/XL2 and DiT-XL/2), and MAR results are reported with a total training of 500 epochs with smaller learning rate (compared to 800 epochs of MAR-H).
$^\dagger$ indicates results with DPM-Solver and 50 steps. }
\label{tab:main_256_full}
\resizebox{0.95\linewidth}{!}{%
\begin{tabular}{@{}lccc|cc|cccc|cccc@{}}
\toprule
\multirow{2}{*}{Gen. Model} &
  \multirow{2}{*}{Tok. Model} &
  \multirow{2}{*}{\# Tokens $\downarrow$} &
  \multirow{2}{*}{Tok. rFID  $\downarrow$} &
  \multirow{2}{*}{Gflops $\downarrow$} &
  \multirow{2}{*}{\begin{tabular}[c]{@{}c@{}}Throughput\\ (imgs/sec)\end{tabular}  $\uparrow$} &
  \multicolumn{4}{c|}{w/o CFG} &
  \multicolumn{4}{c}{w/ CFG} \\ 
 &
   &
   &
   &
   &
   &
  gFID  $\downarrow$ &
  IS  $\uparrow$ &
  Prec.  $\uparrow$ &
  Recall  $\uparrow$ &
  gFID  $\downarrow$ &
  IS $\uparrow$ &
  Prec.  $\uparrow$ &
  Recall  $\uparrow$ 
  \\ \toprule

\multicolumn{14}{c}{\textbf{Auto-regressive Models}} \\ \bottomrule

Taming-Trans.  \cite{esser2021taming} &
  VQ &
  256 &
  7.94 &
  - &
  - &
  5.20 &
  290.3 &
  - &
  - &
  - &
  -  &
  - &
  -
  \\
RQ-Trans.  \cite{lee2022autoregressive} &
  RQ &
  256 &
  3.20 &
  908.91 &
  7.85 &
  3.80 &
  323.7 &
  - &
  - &
  - &
  - &
  - &
  - \\
MaskGIT \cite{chang2022maskgitmaskedgenerativeimage} &
  VQ &
  256 &
  2.28 &
  - &
  - &
  6.18 &
  182.1 &
  0.80 &
  0.51 &
  - &
  -  &
  - &
  - 
  \\
MAGE \cite{li2023magemaskedgenerativeencoder} &
  VQ &
  256 &
  - &
  - &
  - &
  6.93 &
  195.8 &
  - &
 - &
  - &
  -  &
  -  &
  -
  \\
LlamaGen-3B \cite{sun2024autoregressive} &
  VQ &
  256 &
  2.08 &
  781.56 &
  2.90 &
  - &
  - &
  - &
  - &
  3.06 &
  279.7  &
  0.80 &
  0.58 
  \\
TiTok-S-128 \cite{yu2024an} &
  VQ &
  128 &
  1.61 &
  33.03 &
  6.50 &
  - &
  - &
  - &
  - &
  1.97 &
  281.8 &
  - &
  - 
  \\
MAR-H \cite{li2024autoregressiveimagegenerationvector} &
  KL &
  256 &
  1.22 &
  145.08 &
  0.12 &
  2.35 &
  227.8 &
  0.79 &
  0.62 &
  1.55 &
  303.7 & 
  0.81 &
  0.62 
  \\  \midrule
\cellc  &
  \cellc   SoftVQ-L &
\cellc   32 &
\cellc   0.61 &
\cellc   67.93 &
\cellc   2.19 &
\cellc   3.83 &
\cellc   211.2 &
\cellc 0.77  &
\cellc 0.61  &
\cellc   2.54 &
\cellc   273.6  &
\cellc 0.78  &
\cellc 0.61 
\\
\multirow{-2}{*}{\cellc  MAR-H} &
  \cellc  SoftVQ-BL &
\cellc    64 &
\cellc   0.65 &
\cellc   86.55 &
\cellc   0.89 &
\cellc   2.81 &
\cellc   218.3 &
\cellc 0.78  &
\cellc 0.62  &
\cellc   1.93 &
\cellc   289.4    &
\cellc 0.80  &
\cellc 0.61
\\
 \toprule
\multicolumn{14}{c}{\textbf{Diffusion-based Models}} \\ \bottomrule
LDM-4 \cite{rombach2022highresolutionimagesynthesislatent} &
  KL &
  4096 &
  0.27 &
  157.92 &
  0.37 &
  10.56 &
  103.5 &
  0.71 &
  0.62 &
  3.60 &
  247.7 &
  0.87 &
  0.48 \\
U-ViT-H/2$^\dagger$ \cite{bao2023all} &
  \multicolumn{1}{c}{\multirow{5}{*}{KL}} &
  \multirow{5}{*}{1024} &
  \multicolumn{1}{c|}{\multirow{5}{*}{0.62}} &
  128.89 &
  0.98 &
  - &
  - &
  -  &
  - &
  2.29 &
  263.9 & 
  0.82  &
  0.57 
  \\
MDTv2-XL/2 \cite{gao2023mdtv2} &
   &
   &
   &
  125.43 &
  0.59 &
  5.06 &
  155.6 &
  0.72 &
  0.66 &
  1.58 &
  314.7  &
  0.79  &
  0.65
  \\
DiT-XL/2 \cite{peebles2023scalablediffusionmodelstransformers} &
   &
   &
   &
  80.73 &
  0.51 &
  9.62 &
  121.5 &
   0.67 &
  0.67 &
  2.27 &
  278.2   &
  0.83 &
  0.53
  \\
SiT-XL/2 \cite{ma2024sit} &
   &
   &
   &
  \multirow{2}{*}{81.92} &
  \multirow{2}{*}{0.54} &
  8.30 &
  131.7 &
   0.68  &
  0.67 &
  2.06 &
  270.3 &
  0.82 &
  0.59 \\
+ REPA \cite{yu2024representation} &
   &
   &
   &
   &
   &
  5.90 &
  157.8 &
  0.70 &
  0.69 &
  1.42 &
  305.7 &
  0.80 &
  0.65 \\ \midrule
\cellc &
\cellc   SoftVQ-B &
\cellc &
\cellc   0.89 &
\cellc &
\cellc   8.94 &
\cellc   9.83 &
\cellc   113.8 &
\cellc 0.70  &
\cellc 0.61  &
\cellc   3.91  &
\cellc    264.2 &
\cellc  0.81 &
\cellc 0.54  
\\
\cellc  &
\cellc   SoftVQ-BL &
\cellc    &
\cellc   0.68 &
\cellc    &
\cellc   8.81 &
\cellc   9.22 &
\cellc   115.8 &
 \cellc 0.71 &
 \cellc 0.61  &
\cellc   3.78  &
\cellc   266.7  &
\cellc 0.82  &
\cellc 0.54
\\
  \multirow{-3}{*}{\cellc DiT-XL} &
\cellc   SoftVQ-L &
   \multirow{-3}{*}{\cellc  32} &
\cellc    0.74 &
     \multirow{-3}{*}{\cellc 14.52} &
\cellc   8.74 &
\cellc   9.07 &
\cellc   117.2 &
 \cellc 0.71  &
 \cellc 0.61  &
\cellc   3.69 &
\cellc   270.4 &
\cellc 0.83  &
\cellc 0.53  
\\   \midrule
\cellc   &
\cellc   SoftVQ-B &
\cellc   &
\cellc   0.88 &
\cellc   &
\cellc   4.70 &
\cellc   6.62  &
\cellc   129.2 &
 \cellc 0.75 &
\cellc 0.62  &
\cellc   3.29  &
\cellc   262.5 &
\cellc  0.84 &
\cellc 0.54 \\
\cellc   &
\cellc   SoftVQ-BL &
 \cellc   &
\cellc   0.65 &
\cellc    &
\cellc   4.59 &
\cellc   6.53 &
\cellc   131.9 &
\cellc 0.75  &
\cellc 0.62  &
 \cellc  3.11 &
 \cellc   268.3 &
\cellc 0.84  &
\cellc 0.56 \\
\multirow{-3}{*}{\cellc DiT-XL} &
\cellc   SoftVQ-L &
\multirow{-3}{*}{\cellc 64} &
\cellc   0.61 &
  \multirow{-3}{*}{\cellc  28.81} &
\cellc   4.51 &
\cellc   5.83 &
\cellc   141.3 &
\cellc 0.75  &
\cellc 0.62  &
\cellc   2.93  &
\cellc   268.5  &
\cellc  0.84  &
\cellc 0.55
\\ \midrule
\cellc  &
\cellc   SoftVQ-B &
\cellc    &
 \cellc  0.89 &
\cellc    &
\cellc   10.28 &
\cellc   7.99 &
\cellc   129.3 &
\cellc 0.70  &
\cellc 0.63  &
\cellc   2.51 &
\cellc   301.3 & 
\cellc 0.76  &
\cellc 0.62  
\\
\cellc  &
\cellc   SoftVQ-BL &
\cellc    &
\cellc   0.68 &
\cellc    &
\cellc   10.12 &
\cellc   7.67  &
\cellc   133.9 &
\cellc 0.70  &
\cellc 0.63  &
\cellc   2.44 &
\cellc   308.8 &
\cellc 0.76  &
\cellc 0.63  
\\
 \multirow{-3}{*}{\cellc SiT-XL} &
\cellc   SoftVQ-L &
 \multirow{-3}{*}{\cellc   32} &
\cellc  0.74 &
\multirow{-3}{*}{\cellc 14.73}  &
\cellc   10.11 &
\cellc   7.59 &
\cellc   137.4 &
\cellc 0.71  &
\cellc 0.63  &
\cellc   2.44 &
\cellc   310.6 &
\cellc 0.77  &
\cellc 0.63 
\\ \midrule
\cellc   &
\cellc    SoftVQ-B &
\cellc     &
\cellc    0.88 &
\cellc    &
\cellc    5.51 &
\cellc    5.98 &
\cellc    138.0 &
\cellc 0.74  &
\cellc 0.64  &
\cellc    1.78 &
 \cellc   279.0 &
\cellc 0.80  &
\cellc 0.63 
 \\
\cellc &
\cellc    SoftVQ-BL &
\cellc     &
\cellc    0.65 &
 \cellc    &
\cellc    5.42 &
\cellc    5.80 &
\cellc    143.5 &
\cellc 0.74  &
\cellc 0.64  &
\cellc    1.88 &
\cellc    287.9 &
\cellc 0.80 &
\cellc 0.63
\\
 \multirow{-3}{*}{\cellc SiT-XL} &
\cellc    SoftVQ-L &
\multirow{-3}{*}{\cellc 64}  &
\cellc    0.61 &
 \multirow{-3}{*}{\cellc 29.23}  &
\cellc    5.39 &
\cellc    5.35 &
\cellc    151.2 &
\cellc 0.74  &
\cellc 0.64  &
\cellc    1.86 &
\cellc    293.6   &
\cellc 0.81  &
\cellc 0.63
\\ 
 \bottomrule
\end{tabular}%
}
\end{table*}

\begin{table*}[t!]
\centering
\caption{\textbf{System-level comparison} on ImageNet 512$\times$512 conditional generation. We compare with both diffusion-based models and auto-regressive models with different types of tokenizers.  \colorbox{lightgray!20}{Grey} denotes SoftVQ. $^\dagger$ indicates results with DPM-Solver and 50 steps. }
\label{tab:main_512_full}
\resizebox{0.95\linewidth}{!}{%
\begin{tabular}{@{}lccc|cc|cccc|cccc@{}}
\toprule
\multirow{2}{*}{Gen. Model} &
  \multirow{2}{*}{Tok. Model} &
  \multirow{2}{*}{\# Tokens $\downarrow$} &
  \multirow{2}{*}{Tok. rFID  $\downarrow$} &
  \multirow{2}{*}{Gflops $\downarrow$} &
  \multirow{2}{*}{\begin{tabular}[c]{@{}c@{}}Throughput\\ (imgs/sec)\end{tabular}  $\uparrow$} &
  \multicolumn{4}{c|}{w/o CFG} &
  \multicolumn{4}{c}{w/ CFG} \\ 
 &
   &
   &
   &
   &
   &
  gFID  $\downarrow$ &
  IS  $\uparrow$ &
  Prec.  $\uparrow$ &
  Recall  $\uparrow$ &
  gFID  $\downarrow$ &
  IS $\uparrow$  &
  Prec.  $\uparrow$ &
  Recall  $\uparrow$ 
  \\ \toprule

\multicolumn{14}{c}{\textbf{Generative Adversarial Models}} \\ \bottomrule

BigGAN \cite{chang2022maskgitmaskedgenerativeimage} &
  - &
  - &
  - &
  - &
  - &
  - &
  - &
  - &
  - &
  8.43 &
  177.9   &
  - &
  - \\

StyleGAN-XL \cite{karras2019style} &
  - &
  - &
  - &
  - &
  - &
  - &
  - &
  -  &
  - &
  2.41 &
  267.7   &
  - &
  - \\ \bottomrule

\multicolumn{14}{c}{\textbf{Auto-regressive Models}} \\ \toprule

MaskGIT \cite{chang2022maskgitmaskedgenerativeimage} &
  VQ &
  1024 &
  1.97 &
  - &
  - &
  7.32 &
  156.0 &
  -  &
  - &
  - &
  -   &
 -  &
  \\
TiTok-B \cite{yu2024an} &
  VQ &
  128 &
  1.52 &
  - &
  - &
  - &
  - &
  -  &
  - &
  2.13 &
  261.2   &
  - &
  - \\  \midrule 
 \cellc MAR-H &
\cellc  SoftVQ-BL &
  \cellc 64 &
\cellc  0.71 &
\cellc  86.55 &
\cellc  1.50 &
\cellc  8.21 &
\cellc  152.9 &
\cellc 0.69 &
\cellc 0.59 &
\cellc  3.42 &
\cellc  261.8 &
\cellc 0.77 &
\cellc 0.61
\\ \toprule
\multicolumn{14}{c}{\textbf{Diffusion-based Models}} \\ \bottomrule
ADM \cite{dhariwal2021diffusionmodelsbeatgans} &
  - &
  - &
  - &
  - &
  - &
  23.24 &
  58.06 &
  -  &
  - &
  3.85 &
  221.7 &
  0.84 &
  0.53 \\
U-ViT-H/4$^\dagger$ \cite{bao2023all} &
  \multicolumn{1}{c}{\multirow{3}{*}{KL}} &
  \multirow{3}{*}{4096} &
  \multicolumn{1}{c|}{\multirow{3}{*}{0.62}} &
  128.92 &
  0.58 &
  - &
  - &
  -  &
  - &
  4.05 &
  263.8   &
  0.84 &
  0.48 \\

DiT-XL/2 \cite{peebles2023scalablediffusionmodelstransformers} &
   &
   &
   &
  373.34 &
  0.10 &
  9.62 &
  121.5 &
   - &
 -  &
  3.04 &
  240.8   &
  0.84 &
 0.54 \\

SiT-XL/2 \cite{ma2024sit} &
   &
   &
   &
  373.32 &
  0.10 &
  - &
  - &
  -  &
  - &
  2.62 &
  252.2   &
  0.84 &
  0.57 \\  

DiT-XL \cite{chen2024deep} &
  \multicolumn{1}{c}{\multirow{2}{*}{AE}} &
  \multicolumn{1}{c}{\multirow{2}{*}{256}} &
  \multicolumn{1}{c|}{\multirow{2}{*}{0.22}} &
  80.75 &
  1.02 &
  9.56 &
  - &
  -  &
  - &
  2.84 &
  -   &
 - &
 - \\

UViT-H$^\dagger$ \cite{chen2024deep} &
  &
   &
   &
  128.90 &
  6.14 &
  9.83 &
  - &
  -  &
  - &
  2.53 &
  - &
  -  &
  - 
  \\

UViT-H$^\dagger$ \cite{chen2024deep} &
  \multicolumn{1}{c}{\multirow{2}{*}{AE}} &
  \multicolumn{1}{c}{\multirow{2}{*}{64}} &
  \multicolumn{1}{c|}{\multirow{2}{*}{0.22}}  &
  32.99 &
  15.23 &
  12.26 &
  - &
  -  &
  - &
  2.66 &
  -   &
  - &
  - \\

UViT-2B$^\dagger$ \cite{chen2024deep} &
  &
   &
   &
  104.18 &
  9.44 &
  6.50 &
  - &
  -  &
  - &
  2.25 &
  -   &
  - &
  - \\

  \midrule
\cellc & 
\cellc  SoftVQ-L &
\cellc  32 &
\cellc  0.64 &
\cellc  14.73 &
\cellc  10.12 &
\cellc  10.17 &
\cellc  119.2 &
\cellc 0.65 &
\cellc 0.59  &
\cellc 4.23  & 
\cellc 218.0 &
\cellc 0.83 &
\cellc 0.52 
\\
 \multirow{-2}{*}{\cellc SiT-XL} &
\cellc  SoftVQ-BL &
\cellc  64 &
\cellc  0.71 &
\cellc  29.23 &
\cellc  5.38 &
\cellc  7.96 &
\cellc  133.9 &
\cellc  0.73 &
\cellc  0.63 &
\cellc  2.21 &
\cellc  290.5   &
\cellc 0.85  &
\cellc 0.59  
\\
 \bottomrule
\end{tabular}%
}
\end{table*}

We report the full results, including precision and recall, for the ImageNet 256$\times$256 benchmark in \cref{tab:main_256_full} and 512$\times$512 benchmark in \cref{tab:main_512_full}, respectively. 
All main results reported are evaluated on SiT-XL and DiT-XL models trained for 3M iterations, and MAR models trained for 500 epochs.
Noteworthy is that our models achieve performance comparable to state-of-the-art systems.
More importantly, on 512$\times$512 generation, SoftVQ with 32 tokens provides a \textbf{100}x speedup on the inference throughput. 

An interesting finding is that, even though our models in general present the best conditional gFID without using CFG, their performance improvement with CFG becomes smaller, compared to SiT-XL/2 + REPA \cite{yu2024representation}. 
We leave the investigation of the reasons behind this observation for our future work.

We also provide results of longer training, \ie, 4M iterations, and results along the training in \cref{tab:longer-train}.
One can observe that the performance of our models does not saturate at 3M iterations of training, and longer training may provide further performance improvement.

\begin{table*}[t!]
\centering
\caption{Generation performance over training of DiT-XL and SiT-XL trained on SoftVQ-L with 32 and 64 tokens.}
\label{tab:longer-train}
\resizebox{0.6 \linewidth}{!}{%
\begin{tabular}{@{}c|c|cccc|cccc@{}}
\toprule
\multirow{2}{*}{Model}                                                        & \multirow{2}{*}{Training Iter.} & \multicolumn{4}{c}{w/o CFG} & \multicolumn{4}{c}{w/ CFG} \\ \cmidrule(l){3-10} 
 &    & FID & IS & Prec. & Recall & FID & IS & Prec. & Recall \\ \midrule
\multirow{2}{*}{\begin{tabular}[c]{@{}c@{}}DiT/XL\\ SoftVQ-L 32\end{tabular}} & 3M                            &    9.07   &  117.2     & 0.71     &  0.61    & 3.69      &   270.4   &  0.83    &  0.53    \\
 & 4M &      8.54  &  124.0     &  0.72  & 0.62    &   3.58  &  281.9  &  0.82     &    0.53    \\ 
 \midrule
\multirow{2}{*}{\begin{tabular}[c]{@{}c@{}}DiT/XL\\ SoftVQ-L 64\end{tabular}} & 3M                            &   5.83    &   141.3    &   0.75   &   0.62   &   2.93    &  268.5    &   0.84   & 0.55      \\
 & 4M &      5.60  & 144.5      &  0.76 & 0.63      &   2.84  &  270.1  &  0.85     & 0.54       \\
 \midrule
 \multirow{5}{*}{\begin{tabular}[c]{@{}c@{}}SiT/XL\\ SoftVQ-L 32\end{tabular}} & 400K                            &  16.49     &   79.1    &  0.65    &  0.61    &   4.61    &  281.2    & 0.86     & 0.47     \\
 & 1M &  10.30   & 109.0   &  0.69     &     0.63   &  2.85   & 284.0   &   0.76    &  0.61      \\
 & 2M &   8.52  &  122.8  &   0.71    &   0.64     &  2.51   & 296.3   &  0.77     &   0.63     \\
 & 3M &  7.59   & 137.4   &  0.71     &  0.63      &  2.44   & 310.6   &  0.77     & 0.63       \\
 & 4M &     7.32  &    138.6   &  0.72  &  0.63  &  2.38   &   311.5  &  0.77     &    0.63    \\
 \midrule
  \multirow{5}{*}{\begin{tabular}[c]{@{}c@{}}SiT/XL\\ SoftVQ-L 64\end{tabular}} & 400K                    &   10.03     &  103.4     &  0.71     &  0.62    &   3.12   &  236.3     &  0.77    &  0.61       \\
 & 1M &   7.14  &  125.4  &  0.72     &   0.64     &     2.09  & 254.9  &   0.79    &  0.62      \\
 & 2M &    5.88 &  140.6  &    0.73   &   0.64     &     1.89  & 284.4  &   0.80   &  0.63      \\
 & 3M &   5.35  &  151.2  &  0.74     &  0.64      &   1.86  &  293.6  &  0.81     &  0.63      \\
 & 4M &     5.21   & 158.5      &  0.75      & 0.64    &  1.79 &  297.8     &  0.81 & 0.64      \\
 \bottomrule
\end{tabular}%
}
\end{table*}

\subsection{More Comparison to TiTok}

We provide a comparison between TiTok and SoftVQ in \cref{tab:titok_vs_softvq}, including token numbers, model parameters, rFID of IN-1K and MSCOCO, linear probing (LP) accuracy, gFID and IS of SiT-L. 
We show that SoftVQ significantly outperforms TiTok with VQ and KL.
Also, we need to note that while the model architectures are similar, training is different. TiTok is trained using 2-stage decoders with a pre-trained tokenizer, whereas our method is trained end-to-end with a single decoder. 
Also note that, although soft assignment of codewords may not be a new idea, our work is indeed the first to explore it with continuous tokenizer that achieves high compression ratio.

\begin{table}[h!]
\centering
\vspace{-0.1in}
\caption{Comparison between TiTok and SoftVQ.}
\vspace{-0.1in}
\label{tab:titok_vs_softvq}
\resizebox{0.98 \columnwidth}{!}{%
\begin{tabular}{@{}l|cccc|cc@{}}
\toprule
Tokenizer      & \# Param. & rFID (IN-1K) & rFID (COCO) &  LP (IN-1K)  & gFID  & IS    \\ \midrule
TiTok-B-64     & 176       & 1.70 &  11.20 & 39.2  & 25.12    & 53.1     \\ 
TiTok-BL-KL-64 & 389       & 1.25 & 8.94  & 11.4 & 23.35 & 52.7  \\ 
TiTok-BL-VQ-64 & 390       & 2.60 & 11.24 & 9.3 & 19.23 & 61.8  \\ 
SoftVQ-B-64    & \textbf{173}       & \textbf{0.89} & \textbf{5.16} & 42.3  & \textbf{10.13} & \textbf{103.4} \\ \midrule
TiTok-L-32     & 614       & 2.21 & 14.41  & 48.9  &  26.84    & 49.5      \\ 
TiTok-LL-KL-32 & 614       & 1.61 &  10.45 & 12.3 & 24.65 & 51.97 \\ 
SoftVQ-L-32    & \textbf{608}       & \textbf{0.61} & \textbf{5.51} & 59.4 & \textbf{9.47}  & \textbf{107.2} \\ \bottomrule
\end{tabular}%
}
\vspace{-0.1in}
\end{table}

\subsection{Reconstruction Visualization}
\label{sec:appendix-results-recon}

\begin{figure*}
    \centering
    \includegraphics[width=0.95\linewidth]{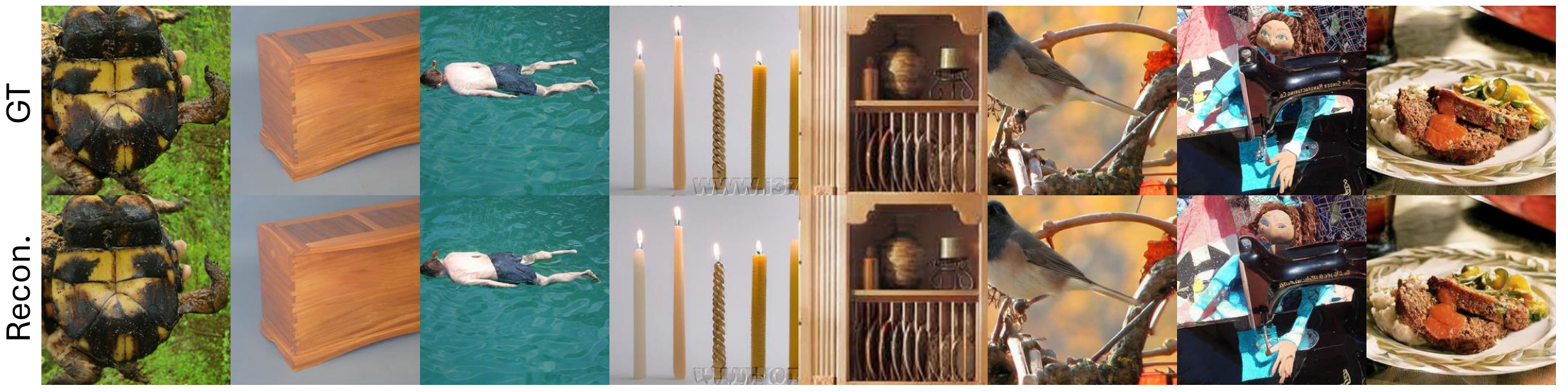}
    \caption{Reconstruction results of SoftVQ-L with 32 tokens.}
    \label{fig:appendix-recon-softvql-32}
\end{figure*}

\begin{figure*}
    \centering
    \includegraphics[width=0.95\linewidth]{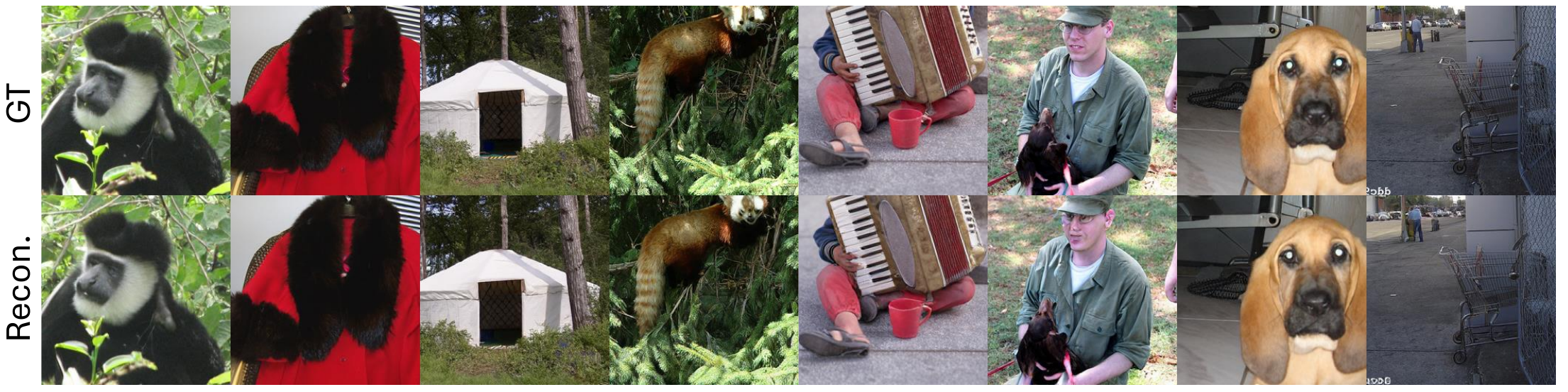}
    \caption{Reconstruction results of SoftVQ-L with 64 tokens.}
    \label{fig:appendix-recon-softvql-64}
\end{figure*}

We present a visualization of the SoftVQ-L reconstruction results with 32 and 64 tokens in \cref{fig:appendix-recon-softvql-32} and \cref{fig:appendix-recon-softvql-64}, respectively.

\subsection{Generation Visualization}
\label{sec:appendix-results-gen}

More visualizations of SiT-XL trained on SoftVQ-L with 32 and 64 tokens are shown here.

\begin{figure*}
    \centering
    \includegraphics[width=0.95\linewidth]{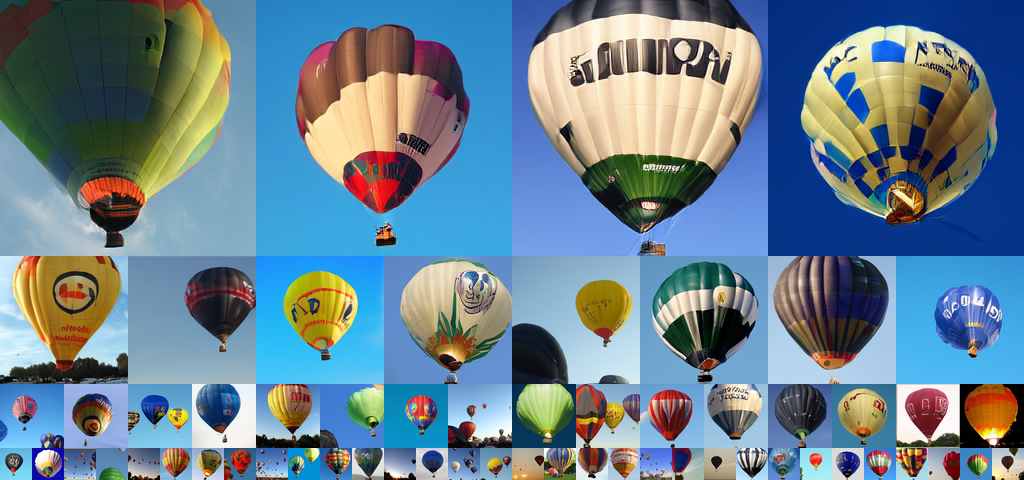}
    \caption{\centering Uncurated 256$\times$256 generation results of DiT-XL with SoftVQ-L 64 tokens. 
    We use CFG with 4.0. Class label = ''balloon'' (417).}
    \label{fig:gen1}
\end{figure*}

\begin{figure*}
    \centering
    \includegraphics[width=0.95\linewidth]{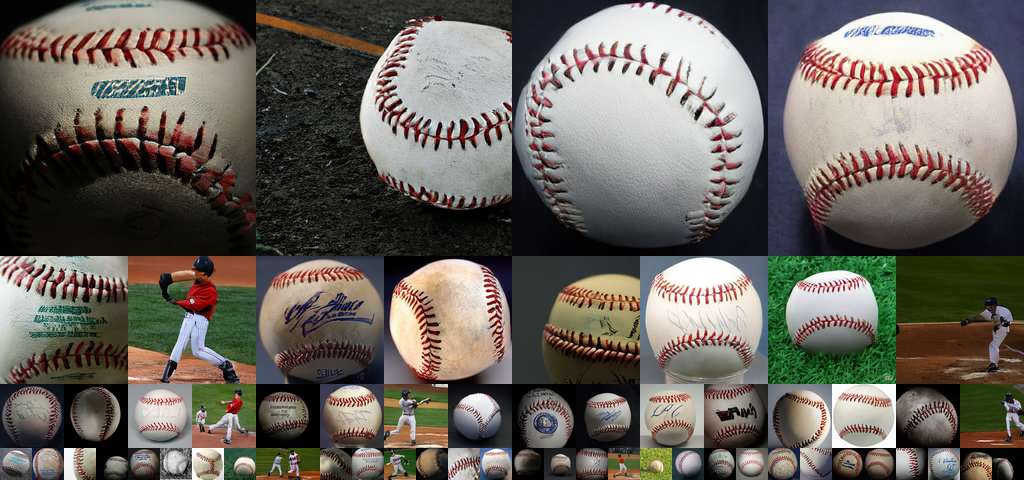}
    \caption{\centering Uncurated 256$\times$256 generation results of DiT-XL with SoftVQ-L 64 tokens. We use CFG with 4.0. Class label = “baseball” (429).}
    \label{fig:gen2}
\end{figure*}

\begin{figure*}
    \centering
    \includegraphics[width=0.95\linewidth]{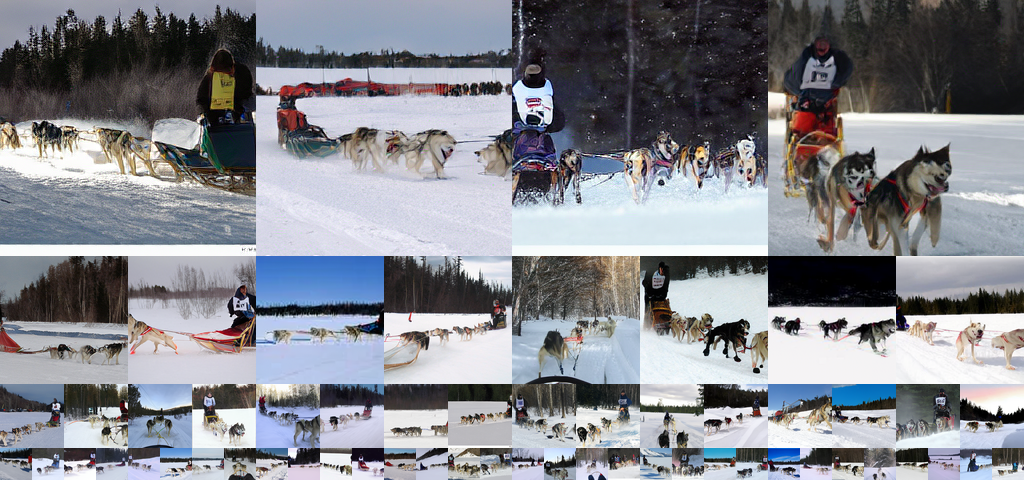}
    \caption{\centering Uncurated 256$\times$256 generation results of DiT-XL with SoftVQ-L 64 tokens. We use CFG with 4.0. Class label = “dog sled” (537).}
    \label{fig:gen3}
\end{figure*}

\begin{figure*}
    \centering
    \includegraphics[width=0.95\linewidth]{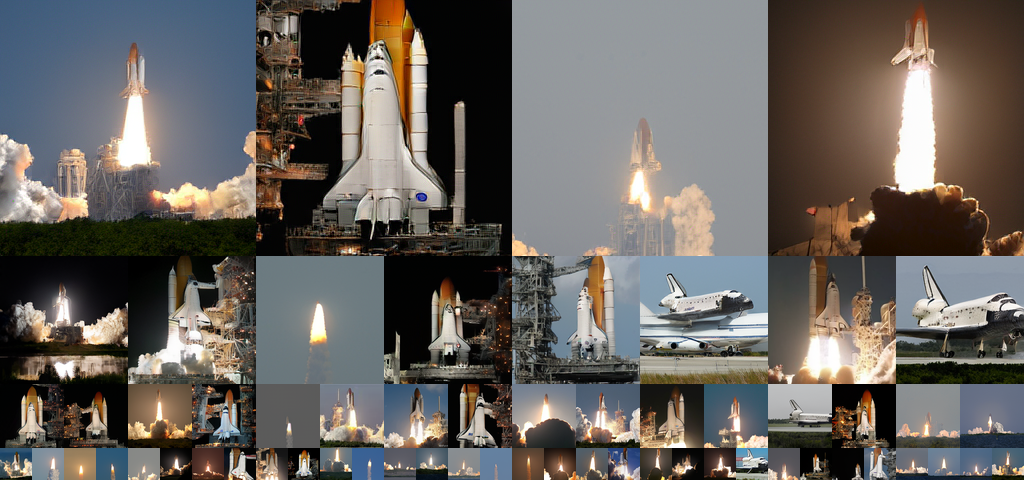}
    \caption{\centering Uncurated 256$\times$256 generation results of DiT-XL with SoftVQ-L 64 tokens. We use CFG with 4.0. Class label = “space shuttle” (812).}
    \label{fig:gen4}
\end{figure*}

\begin{figure*}
    \centering
    \includegraphics[width=0.95\linewidth]{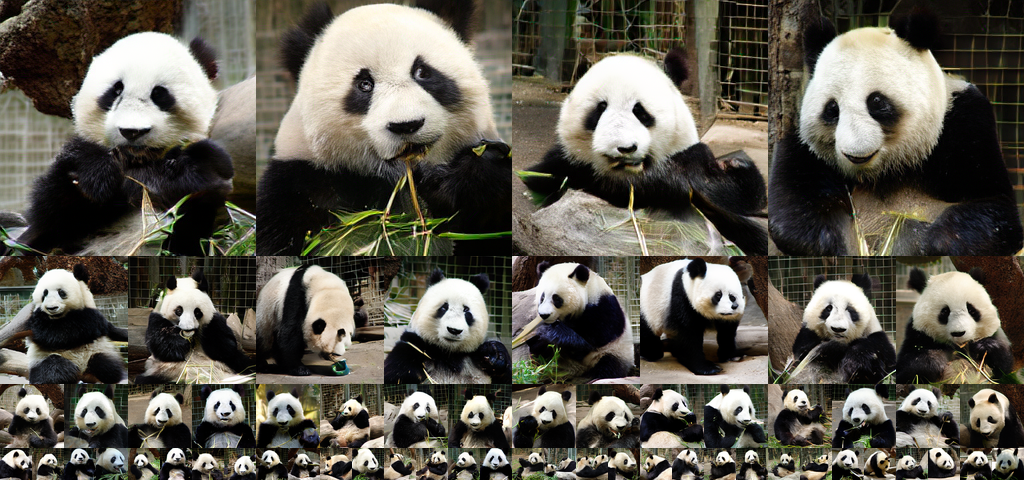}
    \caption{\centering Uncurated 256$\times$256 generation results of DiT-XL with SoftVQ-L 32 tokens. We use CFG with 4.0. Class label = “panda” (388).}
    \label{fig:gen5}
\end{figure*}

\begin{figure*}
    \centering
    \includegraphics[width=0.95\linewidth]{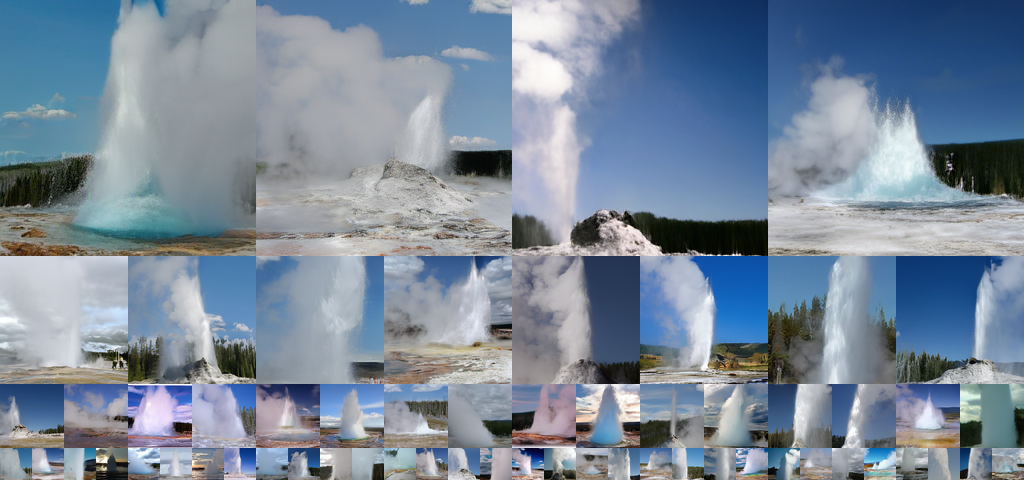}
    \caption{\centering Uncurated 256$\times$256 generation results of DiT-XL with SoftVQ-L 32 tokens. We use CFG with 4.0. Class label = “geyser” (974).}
    \label{fig:gen5}
\end{figure*}

\begin{figure*}
    \centering
    \includegraphics[width=0.95\linewidth]{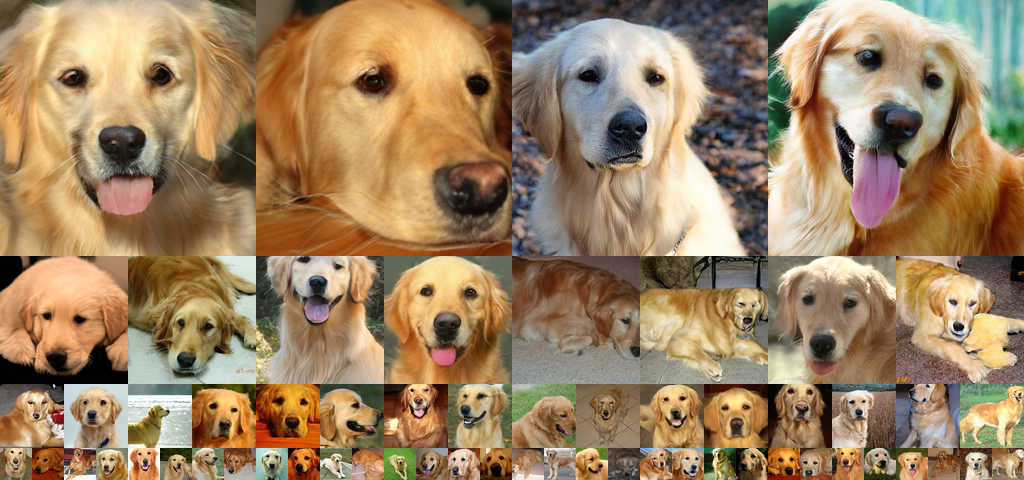}
    \caption{\centering Uncurated 256$\times$256 generation results of SiT-XL with SoftVQ-L 64 tokens. We use CFG with 4.0. Class label = “golden retriever” (207).}
    \label{fig:gen6}
\end{figure*}

\begin{figure*}
    \centering
    \includegraphics[width=0.95\linewidth]{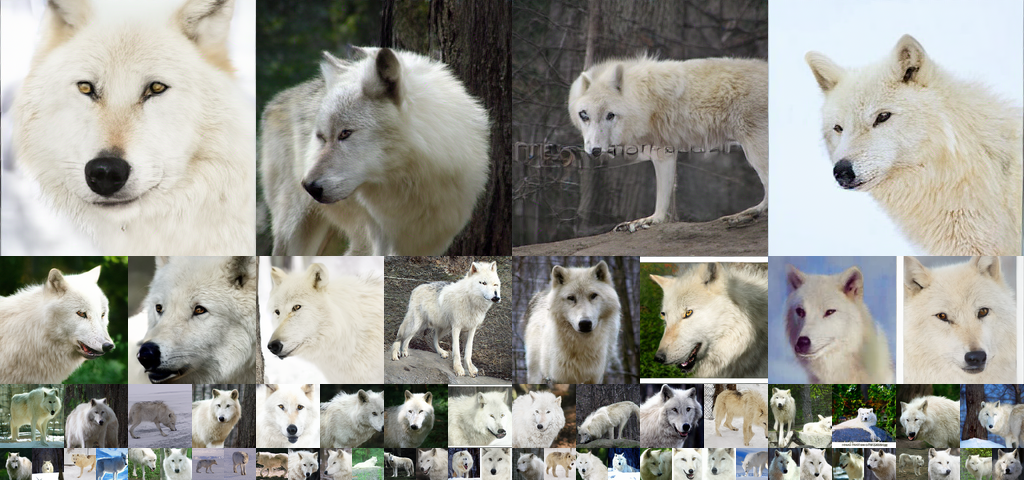}
    \caption{\centering Uncurated 256$\times$256 generation results of SiT-XL with SoftVQ-L 64 tokens. We use CFG with 4.0.  Class label = “arctic wolf” (270).}
    \label{fig:gen6}
\end{figure*}

\begin{figure*}
    \centering
    \includegraphics[width=0.95\linewidth]{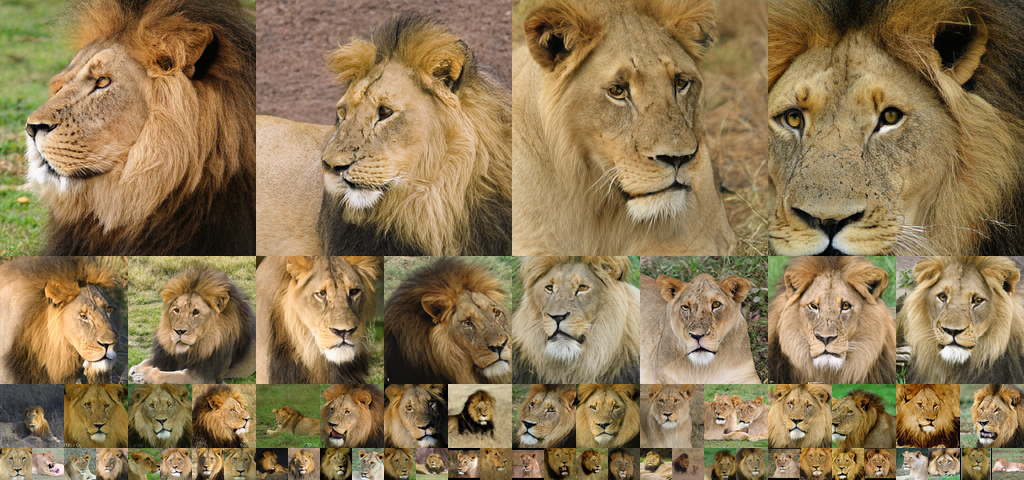}
    \caption{\centering Uncurated 256$\times$256 generation results of SiT-XL with SoftVQ-L 64 tokens. We use CFG with 4.0. Class label = “lion” (291).}
    \label{fig:gen7}
\end{figure*}

\begin{figure*}
    \centering
    \includegraphics[width=0.95\linewidth]{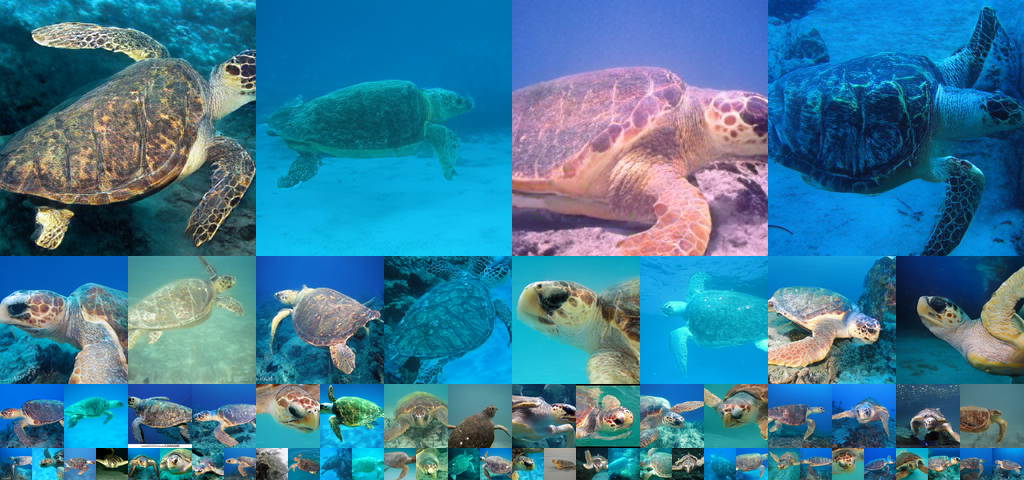}
    \caption{\centering Uncurated 256$\times$256 generation results of SiT-XL with SoftVQ-L 64 tokens. We use CFG with 4.0. Class label = “loggerhead sea turtle” (33).}
    \label{fig:gen8}
\end{figure*}

\begin{figure*}
    \centering
    \includegraphics[width=0.95\linewidth]{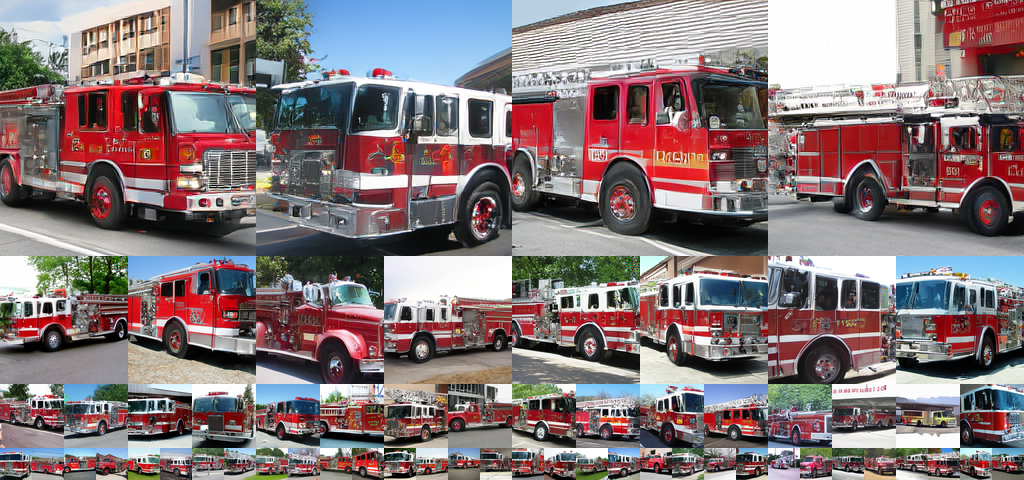}
    \caption{\centering Uncurated 256$\times$256 generation results of SiT-XL with SoftVQ-L 64 tokens.  We use CFG with 4.0. Class label = “fire truck” (555).}
    \label{fig:gen9}
\end{figure*}

\begin{figure*}
    \centering
    \includegraphics[width=0.95\linewidth]{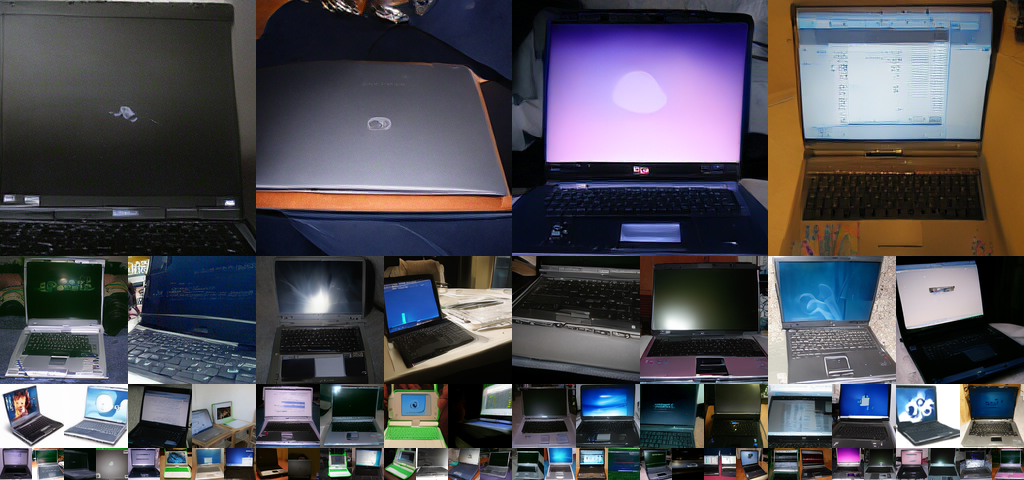}
    \caption{\centering Uncurated 256$\times$256 generation results of SiT-XL with SoftVQ-L 64 tokens. We use CFG with 4.0. Class label = “laptop” (620).}
    \label{fig:gen10}
\end{figure*}

\begin{figure*}
    \centering
    \includegraphics[width=0.95\linewidth]{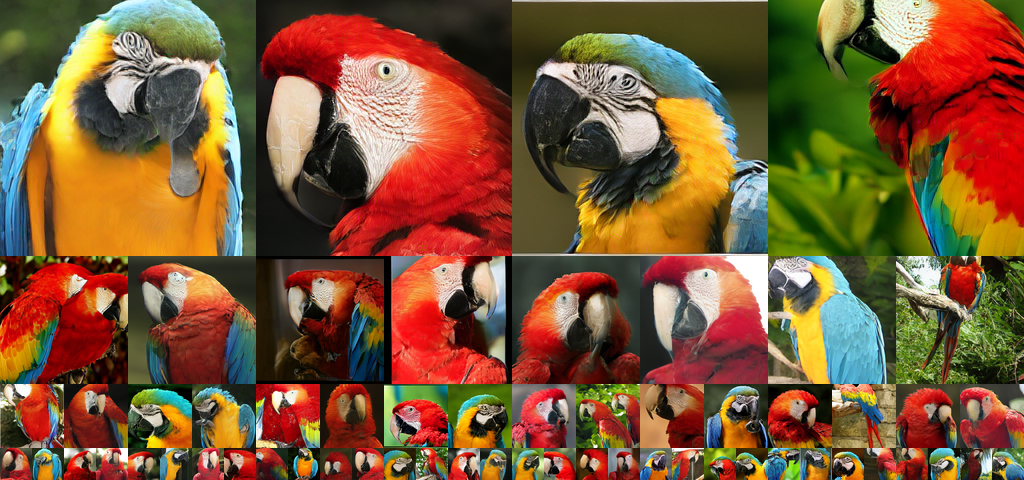}
    \caption{\centering Uncurated 256$\times$256 generation results of SiT-XL with SoftVQ-L 64 tokens. We use CFG with 4.0. Class label = “macaw” (88).}
    \label{fig:gen11}
\end{figure*}

\begin{figure*}
    \centering
    \includegraphics[width=0.95\linewidth]{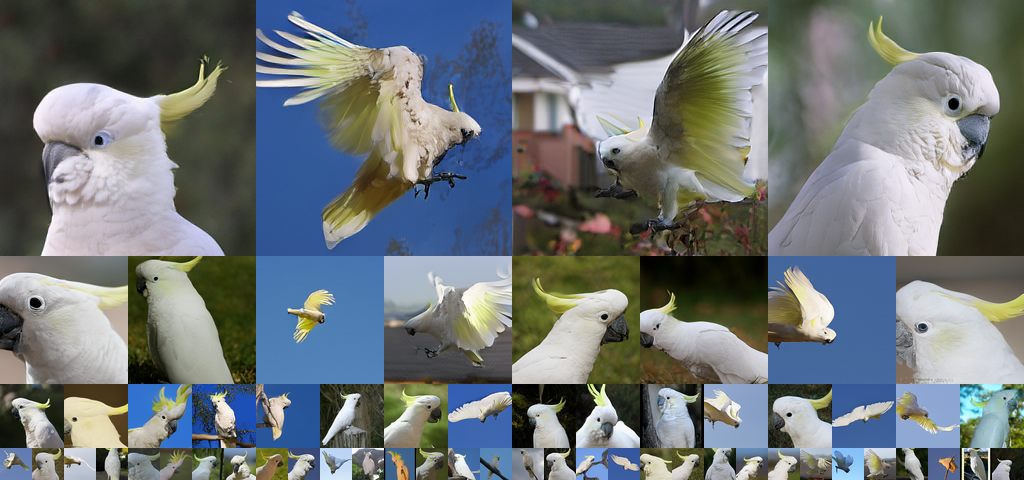}
    \caption{\centering Uncurated 256$\times$256 generation results of SiT-XL with SoftVQ-L 64 tokens. We use CFG with 4.0. Class label = “sulphur-crested cockatoo” (89).}
    \label{fig:gen12}
\end{figure*}

\begin{figure*}
    \centering
    \includegraphics[width=0.95\linewidth]{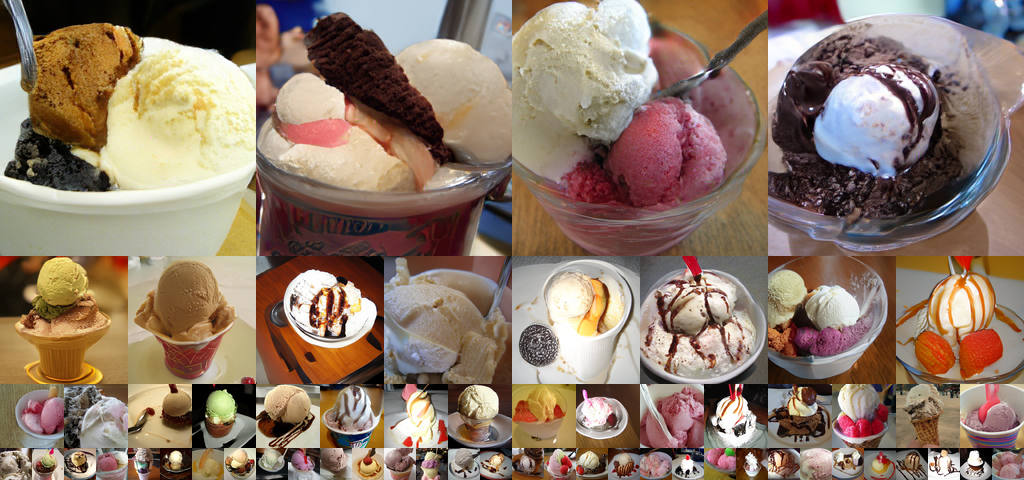}
    \caption{\centering Uncurated 256$\times$256 generation results of SiT-XL with SoftVQ-L 64 tokens. We use CFG with 4.0. Class label = “ice cream” (928).}
    \label{fig:gen13}
\end{figure*}

\begin{figure*}
    \centering
    \includegraphics[width=0.95\linewidth]{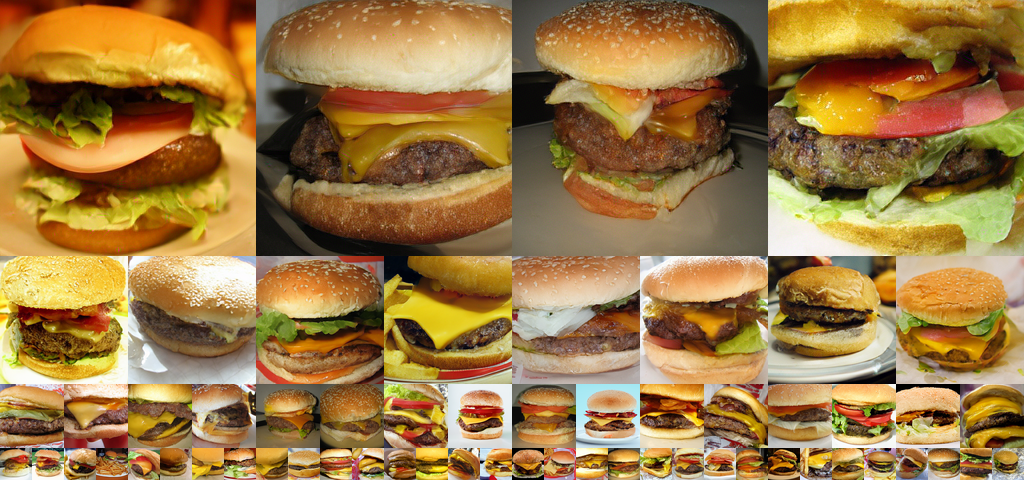}
    \caption{\centering Uncurated 256$\times$256 generation results of SiT-XL with SoftVQ-L 64 tokens. We use CFG with 4.0. Class label = “cheeseburger” (933).}
    \label{fig:gen14}
\end{figure*}

\begin{figure*}
    \centering
    \includegraphics[width=0.95\linewidth]{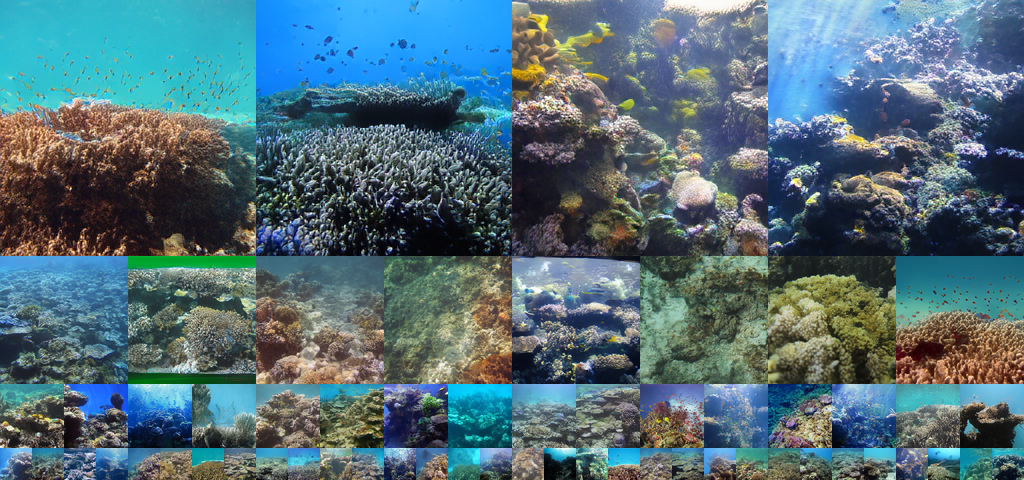}
    \caption{\centering Uncurated 256$\times$256 generation results of SiT-XL with SoftVQ-L 32 tokens.  We use CFG with 4.0. Class label = “coral reef” (973).}
    \label{fig:gen15}
\end{figure*}

\begin{figure*}
    \centering
    \includegraphics[width=0.95\linewidth]{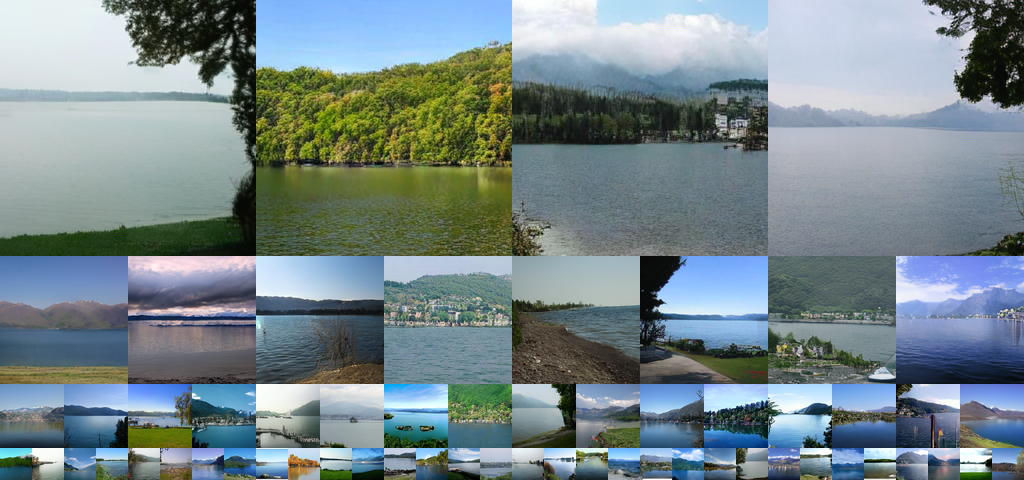}
    \caption{\centering Uncurated 256$\times$256 generation results of SiT-XL with SoftVQ-L 32 tokens. We use CFG with 4.0. Class label = “lake shore” (975).}
    \label{fig:gen16}
\end{figure*}

\begin{figure*}
    \centering
    \includegraphics[width=0.95\linewidth]{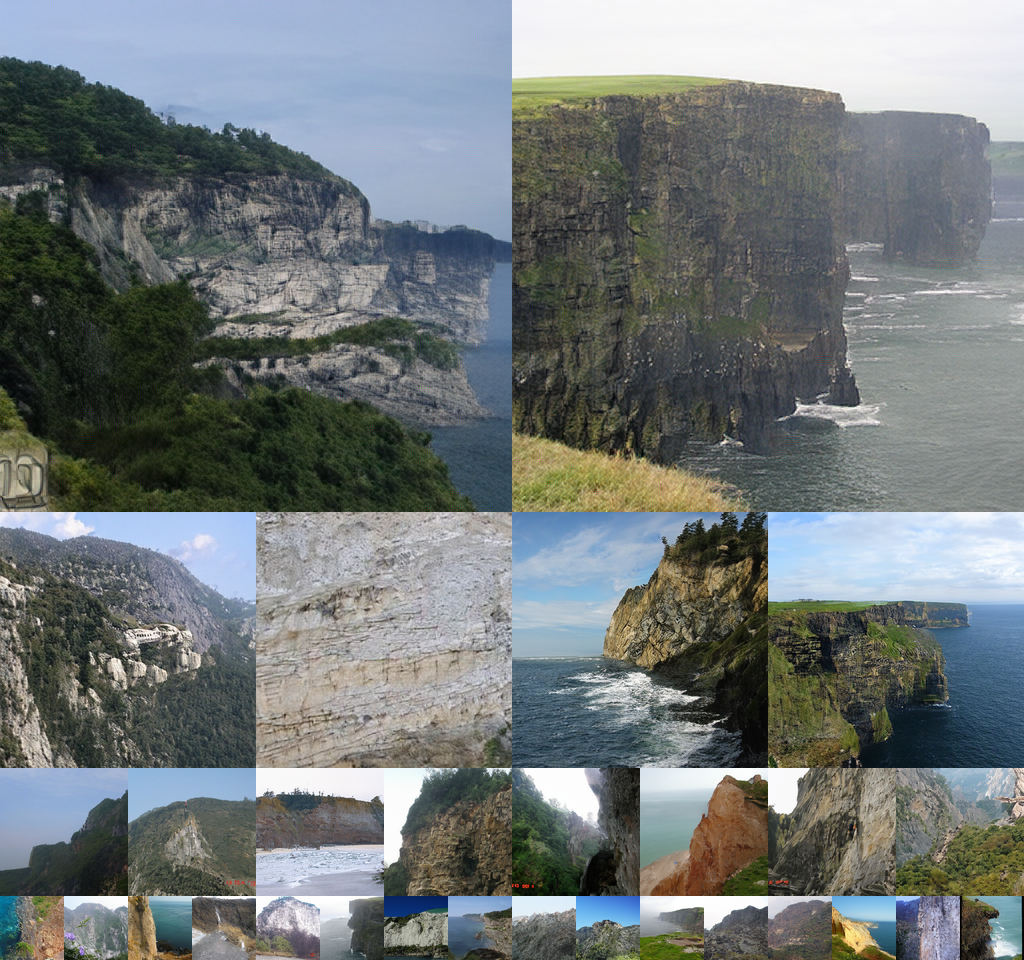}
    \caption{\centering Uncurated 512$\times$512 generation results of SiT-XL with SoftVQ-L 64 tokens.  We use CFG with 2.0. Class label = “cliff drop-off” (972).}
    \label{fig:gen18}
\end{figure*}

\begin{figure*}
    \centering
    \includegraphics[width=0.95\linewidth]{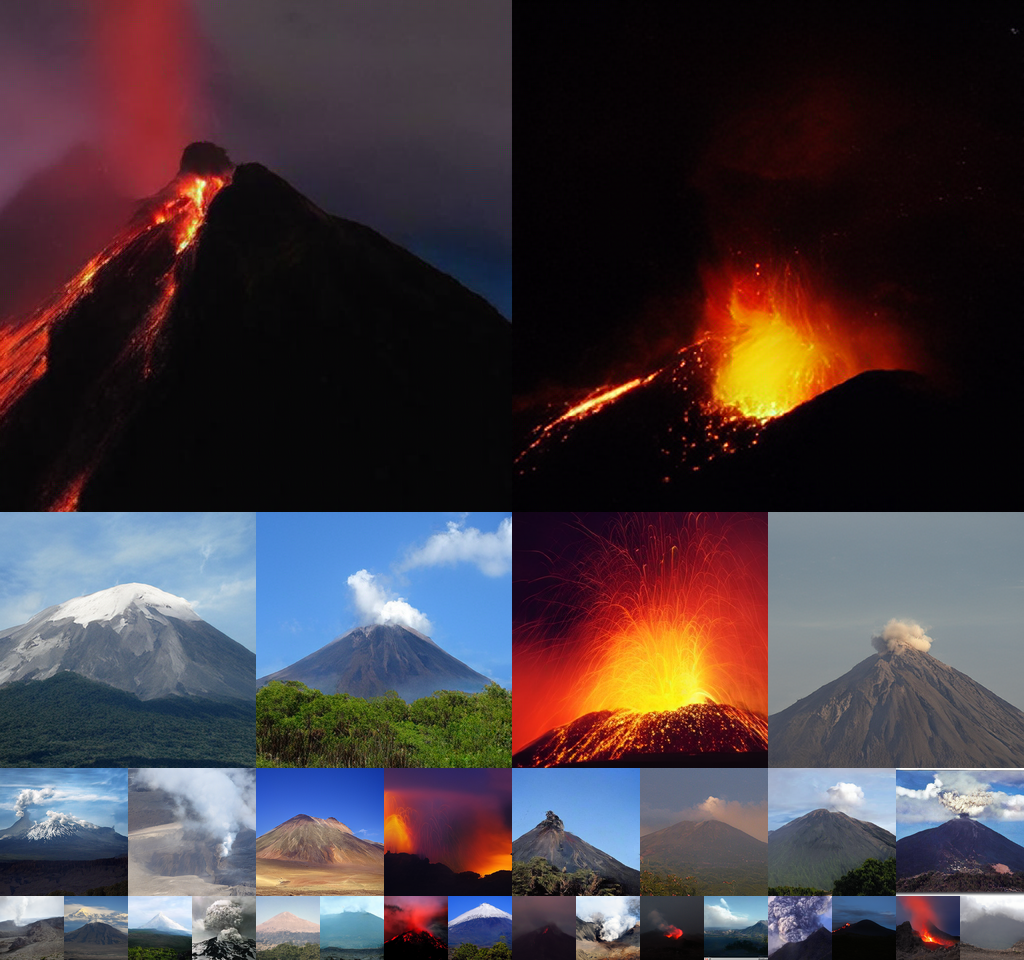}
    \caption{\centering Uncurated 512$\times$512 generation results of SiT-XL with SoftVQ-L 64 tokens.  We use CFG with 2.0. Class label = “volcano” (980).}
    \label{fig:gen17}
\end{figure*}

\end{document}